\documentclass{article}
\usepackage[margin=0.75in, bottom=0.5in]{geometry}
\PassOptionsToPackage{hyphens}{url}  

\usepackage{mathrsfs,amsthm,graphicx,xcolor,verbatim,bbm,amsmath,amsfonts,dsfont,amssymb,newclude,graphicx,enumitem,bm,mathtools}

\usepackage{tikz}
\usetikzlibrary{arrows.meta}
\usepackage{amsmath}
\tikzset{
  bowl/.style = {line width=1.2pt, draw=gray!70},
  etaLess/.style = {-{Stealth[length=3.2pt]}, line width=1pt, draw=teal!70!black},
  etaMore/.style = {-{Stealth[length=3.2pt]}, line width=1pt, draw=orange!80!black},
}

\usepackage{fullpage}

\usepackage{pgfplots}
\usepackage{xcolor}
\definecolor{PythonGreen}{HTML}{2ca02c}
\usepackage{url}            
\usepackage{nicefrac}       
\usepackage{caption}
\usepackage{float}
\usepackage{mdframed}
\usepackage{fontawesome5}
\usepackage{wrapfig} 
\usepackage[T1]{fontenc}    
\usepackage[utf8]{inputenc}

\usepackage{subcaption}
\usepackage{overpic}

\usepackage{hyperref}
\PassOptionsToPackage{numbers,compress}{natbib}
\usepackage{natbib}

\hypersetup{colorlinks=true, linktocpage}

\hypersetup{
    colorlinks,
    linkcolor={red!50!black},
    citecolor={blue!80!black},
    urlcolor={blue!80!black}
}

\theoremstyle{plain}

\theoremstyle{remark}

\theoremstyle{definition}

\newtheorem*{theorem*}{Theorem}
\newtheorem*{lemma*}{Lemma}
\newtheorem*{proposition*}{Proposition}
\newtheorem*{definition*}{Definition}

\usepackage[most]{tcolorbox}

\newtcolorbox{summarybox}{
  colframe=black!70,         
  boxrule=2pt,               
  arc=4pt,                   
  left=2pt,
  right=2pt,
  top=3pt,
  bottom=3pt,
  halign=center,
}

\usetikzlibrary{positioning, shapes.geometric, fit, backgrounds, calc}

\usepackage{array}       
\newcolumntype{L}[1]{>{\raggedright\arraybackslash}p{#1}}  
\usepackage{placeins}     
\title{\textbf{pAI\texttt{/MSc}: ML Theory Research\\with Humans \textit{on} the Loop}\\
{\Large A Technical Report}}

\author{%
  Mahmoud Abdelmoneum%
  \thanks{Equal contribution. Correspondence: \href{mailto:pierb@mit.edu}{\texttt{pierb@mit.edu}} or \href{https://discord.gg/Pz7spPPY}{\texttt{Discord}}.
  Please check \href{https://PoggioAI.github.io}{our website for updates}.
  Code available at
  \href{https://github.com/PoggioAI/PoggioAI_MSc}{\texttt{github.com/PoggioAI/PoggioAI\_MSc}}. Please note that we also provide a \texttt{Claude Code Skill} \href{https://github.com/PoggioAI/PoggioAI_MSc-claude}{\texttt{PoggioAI\_MSc-claude}}, this however, produces lower quality manuscripts.
  }$^{\ , {\color{red!50!black}1}, {\color{red!50!black}2}}$, \quad
  Pierfrancesco~Beneventano$^{*,{\color{red!50!black}1}}$, \quad
  Tomaso Poggio$^{{\color{red!50!black}1}}$\\[0.4cm]
{\color{red!50!black}$^1$}\textit{Massachusetts Institute of Technology}\\
{\color{red!50!black}$^2$}\textit{Perseus Labs}
}

\date{V1: March 21, 2026. \quad Current: March 25, 2026}

\begin{document}

\maketitle

\begin{abstract}
    We present \textbf{pAI}\texttt{/MSc}\footnote[3]
    {Which stands for \textbf{\underline{p}}rincipal \textbf{\underline{A}}gentic \textbf{\underline{I}}nvestigator.}, an open-source, customizable, modular multi-agent system for academic research workflows.
    Our goal is not autonomous scientific ideation, nor fully automated research. It is narrower and more practical: to reduce by orders of magnitude the human steering required to turn a specified hypothesis into a literature-grounded, mathematically established, experimentally supported, submission-oriented manuscript draft.
    \textbf{pAI}\texttt{/MSc} is built with a current emphasis on machine learning theory and adjacent quantitative fields.
\end{abstract}

\section{Introduction}
LLMs first, agentic systems now, are changing how research is conducted, including in academia.
Reports of AI systems making progress on difficult mathematical problems 
\citep{funsearchNature2024,funsearchRepo,alphaevolve2025,mathexplore2025,alphaevolveProblemRepo,ramseyAlphaEvolve2026,claudesCycles2026,taoErdos1026_2025,firstProof2026,firstBatchSite2026,openaiFirstProofSubmissions2026,zhangMaFirstProofZenodo2026,openaiGPT52ScienceMath2025,sellkeYinLearningCurve2025,mathIncGauss,mathIncStrongPNTSite,mathIncStrongPNTRepo,fieldsGaussPNTTalk2025,sothanaphanErdos728_2026,harmonicErdos728Post2026,bloomErdos728_2026,bloomErdos729_2026,bloomErdos397_2026,erdosProblemDatabaseRepo,gigazineOpenAIErdosClaims2025,brightOpenAIErdosClaims2025}
naturally force us---\textit{mathematicians} and \textit{computer scientists}---to ask unpleasant and controversial questions:
\begin{center}
    \textit{How much of academic research can, in fact, be automated?}
    \\[0.2cm]
    \textit{To what extent, and how soon, might some of our skills be displaced by such systems? Which ones?}
\end{center}
The \textit{scientific} goal of this effort is to understand to what extent agentic systems can automate \textbf{\textit{high-quality}} research.
The long-term \textit{engineering} ambition of this line of work is to identify, approach, and if possible attain this upper bound of automation compatible with high-quality research.

As of February 2026, we believe current agentic systems can be impressive, but the quality of the ideas and artifacts they produce still falls short of serious academic standards and for high-impact research. Our focus, thus, is developing the best possible system for what we see as a \textbf{key necessary step toward that goal.}
\vspace{0.7cm}

\subsection*{Practical Objective} 
A natural dream for many researchers is to have a junior collaborator who, given the core idea and limited guidance, can turn that idea into something as close as possible to a submission-ready article.
In our experience, moving from a strong human-developed idea to a solid machine-learning-theory manuscript still often takes on the order of $10^3$ (or $5 \cdot 10^4$, the baseline defined by \citep{schwartzVibePhysics2026} to make \citep{schwartzCParameter2026} with Claude) prompts---if possible at all---with frontier reasoning models and agentic systems as Claude Code or GPT Pro.
Perhaps some of this is a prompting problem on our side. But to the best of our knowledge, current AI systems are not yet enough to satisfy this naive dream. 
The goal here is:
\begin{center}
    \textit{Can we design an AI system that, in at most 10 human steers,
    \\[0.2cm] pushes a strong \textbf{Hypothesis} to a \textbf{Written Article} of serious academic quality?}
\end{center}
More precisely, can we build an agentic system that requires no more than 10 explicit interventions from the human researcher to develop an idea rigorously, establish it at the highest academic standards, and produce a strong manuscript draft?
Equivalently, one can view the problem as one of compression: can we reduce the human control budget from roughly $\sim 10^6$ tokens (or more, like $36M$ where used as input tokens in the baseline defined by \citep{schwartzVibePhysics2026}) of interactive instruction to something closer to $\sim 10^3$, without losing rigor and quality?

Importantly, this project is \textit{not}\footnote{Yet.} about fully automating the path from idea to article, \textit{nor} is it about having AI systems originate the research idea itself\footnote{Yet.}:
\begin{center}
    \textit{The key constraint is \textbf{high quality}, the objective function \\[0.2cm] to minimize is the \textbf{human steering burden} (or oversight) required to reach that.}
\end{center}

\subsection*{Researching Ethically: AI Systems Leverage}
A growing literature argues that machine learning scholarship often states central claims more strongly than they are mathematically, causally, or empirically established \citep{liptonSteinhardt2019Troubling,gelman2019TroublingBlog,pineauEtAl2021ImprovingReproducibility,beanEtAl2025MeasuringWhatMatters,oxfordInternetInstitute2025BenchmarkPress}. Direct critiques point to explanation–speculation slippage, “mathiness,” and unclear attribution of empirical gains; reproducibility reforms explicitly identify over-claiming and hypothesis–claim mismatch as part of the field’s standards problem; and domain-specific audits have found systematic overoptimism driven by weak baselines, reporting bias, shortcut learning, and fragile benchmark design \citep{liptonSteinhardt2019Troubling,sculleyEtAl2015HiddenTechnicalDebt,hendersonEtAl2018DeepRLMatters,mcgreivyHakim2024WeakBaselines,geirhosEtAl2020ShortcutLearning,ulmerHardmeierFrellsen2022DeepSignificance}. In causal settings, additional work warns that predictive performance is frequently insufficient to justify intervention or fairness claims \citep{jonesEtAl2024CausalPerspective,broadbentGrote2022CanRobotsDoEpidemiology,collinsMoons2019ReportingAIPredictionModels,fullerWright2024AISnakeOil}.

Producing only rigorously established work is not a methodological preference---it is an \emph{ethics of research}. If researchers will increasingly rely on AI systems, then the highest-leverage place to enforce that ethic is \emph{inside those systems}: bias the tools toward establishing results solidly, and establishment scales with adoption.

\textbf{pAI}\texttt{/MSc} is our attempt to do exactly this. System prompts and control flows are designed to make novelty claims explicit, stress-test them against competing explanations, align mathematics with experiments, and demand intervention experiments when causal language is involved. A field with more papers is not bad per se---provided those papers push the knowledge boundary. We believe optimizing for quality the AI tools that researchers use is the intervention most likely to move the needle.



\subsection*{Why Academic Research Is Harder Than Current Agentic Search}
\textit{How is our setting different from \texttt{AutoResearch} \citep{introKarpathyAutoResearch2026} and related systems?} Not mainly because we target theory, but because academic research offers much weaker objective feedback. Many recent agentic successes depend on three favorable conditions:
\begin{enumerate}[itemsep=0.15em,topsep=0.15em,parsep=0pt]
    \item \textbf{Bounded domain.} AutoResearch edits a single file, \texttt{train.py}.
    \item \textbf{Short evaluation loops.} Each AutoResearch run is capped at five minutes.
    \item \textbf{Automatic objectives.} AutoResearch optimizes validation bits-per-byte; MLAgentBench \citep{introHuangEtAlMLAgentBench2024} likewise studies executable ML tasks with clear goals and automatic evaluation.
\end{enumerate}

In more open-ended settings, evaluation itself becomes part of the problem. The AI Scientist \citep{introLuEtAlAIScientist2024}, as \textbf{pAI}\texttt{/MSc}, relies on simulated review and an automated reviewer; Agent Laboratory \citep{introSchmidgallEtAlAgentLaboratory2025} allows human guidance and finds that it improves research quality; and MLR-Bench  \citep{introChenEtAlMLRBench2025} uses an LLM judge with structured rubrics.

Academic research, especially theory, weakens all three assumptions. The search space is open-ended, significance may become clear only years later, as in delayed recognition and ``sleeping beauties'' \citep{introKeEtAlSleepingBeauties2015}, and no short-horizon scalar objective faithfully captures novelty, correctness, and long-run scientific value. More broadly, the research-assessment literature argues that metrics should inform rather than replace expert judgment, and that target metrics are easy to game \citep{introHicksEtAlLeiden2015,introWilsdonEtAlMetricTide2015,introFireGuestrinGoodhart2019}.
So academic research is only partly amenable to direct metric optimization. Local correctness can often be checked, but end-to-end research value cannot yet be scored reliably by a single automatic reward. That is why our system is organized around artifacts, adversarial critique, and human-on-the-loop judgment rather than one scalar score. A central open problem is to design better computable proxies for research quality without reducing research to the proxy.

\subsection*{Contributions}
\begin{itemize}[itemsep=0.15em,topsep=0.15em,parsep=0pt]
\vspace{-0.1cm}
    \item We present \textbf{pAI}\texttt{/MSc}, an open-source, customizable, modular multi-agent system for hypothesis-to-paper workflows with explicit human oversight.
    \item We introduce an artifact contract in which progress is represented by named intermediate outputs and stage-level validation gates rather than by free-form dialogue alone.
    \item We implement a fixed workflow that decomposes work into 23 specialized agents spanning literature review, theory, experimentation, synthesis, and editorial stages, with repair and follow-up routes when artifacts are incomplete, inconsistent, not novel, or not adequately established.
    \item We integrate optional rigor-enhancing components---including theorem-oriented reasoning, multi-model debate counsel, and tree-search-based exploration---within the same execution framework.
    \item We provide a technical report on the scope, architecture, and operational limits of the current release, with an explicit separation between structural success and scientific validity.
\end{itemize}



\subsection*{Acknowledgements}
We are extremely thankful to Perseus Labs who meaningfully helped both economically and in terms of software development, a particular thank you to Hilal Hussain. We also want to thank deeply Riccardo Neumarker, Emanuele Rimoldi, Yulu Gan, Qianli Liao, Marc Bacvanski, Liu Ziyin, and Theodoros Evgeniou for many fundamental discussions. A deep thank to DeepMind, in particular to Demis Hassabis, for offering API credits for testing earlier versions of the system. \textit{We welcome contributions and are seeking support for the API costs of further development.}

\newpage

\tableofcontents

\newpage


\section{Engineering Lessons and Design Requirements}
\label{sec:requirements}

This section has two goals. First, it makes explicit the main engineering difficulties we encountered while building \textbf{pAI}\texttt{/MSc}, so that future work can attack them directly rather than rediscover them piecemeal. Second, it serves as guidance for users who may customize or extend the system: it documents which design choices were responses to concrete workflow failures, which problems we believe are only partially solved, and which interface contracts should be preserved if one wants the pipeline to remain reliable. We present these lessons in the same order in which they typically appear during a research-to-paper run.

\subsection{Difficulties Encountered}
\paragraph{1. Ideation quality improved when planning became debate rather than monologue.}
One of our earliest observations was that single-agent ideation tended to converge too quickly to generic or aesthetically plausible project plans. Quality improved substantially when the early planning phase was reformulated as a structured debate among multiple agents rather than a single agent trying to ideate, justify, and plan everything at once. This was also oserved, but implemented differently by \citet{du2024improving}.

\paragraph{2. Give agents explicit, competing objectives.}
A second related observation was that debate alone was not enough: each agent needed an explicit goal. In practice, the quality of the resulting research plans increased when we assigned different roles and incentives to different ideation agents. In the current design, these roles include one agent pushing for timely and practitioner-relevant questions, one pushing for rigorous and potentially novel mathematical theory, and one pushing for the strongest narrative arc together with the right alignment or disalignment with existing folklore. This does not solve ideation in general, but it consistently produced better decompositions and more informative disagreements than a single planner.

\paragraph{3. Theory and experiments should be coordinated, but not fully entangled.}
A natural design instinct is to maximize context sharing between the mathematical and experimental tracks. In our experience, that improved local alignment but weakened both tracks. When theory and experiments shared too much context too early, they tended to collapse toward the same framing, inherit each other's blind spots, and produce intermediate outputs that were better matched but less strong. Better final papers came from a weaker form of coordination: both tracks share an explicit decomposition and common contribution claims, but do not continuously share unrestricted working context. Instead, they exchange information at controlled synchronization points, especially before synthesis and at the beginning of a new cycle. This design tries to preserve independence of reasoning while still forcing the two tracks to remain on a single paper trajectory.

\paragraph{4. Long-horizon runs need explicit stopping criteria.}
Earlier versions of the system often kept iterating long after the core artifacts were already sufficiently established. This led to wasted computation, wasted budget, and occasionally worse outputs due to late over-editing or scope drift. The underlying difficulty is that the relevant notion of diminishing returns is conceptual rather than merely token-based: a run can continue to generate text while adding little scientific value. Our current response is to expose explicit stage gates, bounded iteration, and steering checkpoints. When the system judges that additional loops are likely to bring only marginal conceptual gain, it is encouraged to stop, surface the current state, and request a human steer. We do not regard this problem as fully solved, but treating stopping as a first-class systems problem already improves both cost and reliability.

\paragraph{5. Parallel debugging remains expensive and only partially solved.}
Debugging a long, partially parallel research workflow is much slower than debugging a single-turn assistant. Failures may arise from malformed artifacts, routing errors, stale intermediate state, synchronization bugs between tracks, or interactions between multiple agents that only become visible late in the run. When these failures occur after several hours of execution, the cost of diagnosis can be substantial. We do not currently know how to make this problem easy. What we do instead is make failure state external and inspectable: the system persists checkpoints, intermediate messages, budget traces, and mode-specific workspaces so that a run can be resumed, audited, or post-mortemed rather than restarted from scratch. This does not remove the debugging burden, but it makes it more manageable and far less opaque.

\subsection{Implications on Workflow-Level Design Requirements}
Taken together, the observations above led us to treat workflow design itself as a first-class systems problem. In our experience, long research runs become unreliable when intermediate state is implicit, when theory and experiments evolve without a shared contract, or when apparently polished drafts are produced without the artifacts needed to inspect how those drafts were assembled. The current release therefore follows seven design requirements.

\begin{itemize}
    \item \textbf{Reduce steering burden without hiding human responsibility.} The aim is to compress the amount of human orchestration needed to move from an initial research objective to a paper draft. At the same time, novelty, correctness, citation faithfulness, and publication decisions remain human responsibilities.
    
    \item \textbf{Keep theory and experiments on a single paper trajectory.} When both tracks are active, they must remain tied to the same contribution claims and follow-up logic, even when they are developed semi-independently.
    
    \item \textbf{Externalize intermediate state as named artifacts.} Progress should live in files and structured outputs rather than only in conversational context, so that runs are inspectable, resumable, and auditable.
    
    \item \textbf{Support bounded iteration rather than a single forward pass.} Real research workflows need loopbacks: feasibility checks may fail, experiments may invalidate theory, theory may require new experiments, and writing may reveal unresolved inconsistencies.
    
    \item \textbf{Separate structural validation from scientific truth.} The system can validate artifact existence, parseability, selected internal consistency properties, and review-gate thresholds. It cannot certify scientific correctness, novelty, or acceptance probability.
    
    \item \textbf{Make long runs resumable, steerable, and budget-aware.} Checkpointing, stage-based resume, live steering hooks, and budget traces are operational requirements rather than convenience features.
    
    \item \textbf{Expose multiple enforcement levels.} The same architecture should support lighter exploratory runs and stricter paper-oriented runs, with different validation thresholds and artifact expectations.
\end{itemize}

\subsection{Artifact Contracts}
A run is not treated as successful merely because it produces fluent prose. It is expected to produce a paper-centered workspace with explicit deliverables for discovery, planning, execution, synthesis, and editorial review. This artifact contract is the main interface between the workflow and the researcher. It serves both goals of the present section: it tells future researchers which interface stabilized the pipeline, and it warns customizers which artifacts are dangerous to remove if they want the system to remain inspectable, resumable, and debuggable.

\begin{table}[t]
\centering
\footnotesize
\setlength{\tabcolsep}{4pt}
\begin{tabular}{|L{2.0cm}|L{5.4cm}|L{3.8cm}|L{2.6cm}|}
\hline
\textbf{Bundle} & \textbf{Concrete artifacts} & \textbf{Role} & \textbf{Gate / mode} \\
\hline
Paper workspace core &
  \texttt{final\_paper.tex},
  \texttt{paper\_workspace/\allowbreak literature\_review.pdf},
  \texttt{research\_plan.pdf},
  \texttt{results\_assessment.pdf},
  \texttt{followup\_decision.json},
  \texttt{track\_decomposition.json}
& Minimal paper-facing workspace linking discovery, planning, synthesis, and manuscript generation
& \texttt{--enforce-paper-\allowbreak artifacts} \\
\hline
Optional outputs &
  \texttt{final\_paper.pdf},
  \texttt{experiments\_to\_run\_later.md}
& Tighten success definition for PDF or deferred experiment planning
& \texttt{--require-pdf}; \texttt{--require-experiment-\allowbreak plan} \\
\hline
Editorial strictness &
  \texttt{author\_style\_guide.md},
  \texttt{intro\_skeleton.tex},
  \texttt{style\_macros.tex},
  \texttt{reader\_contract.json},
  \texttt{editorial\_contract.md},
  \texttt{theorem\_map.json},
  \texttt{revision\_log.md},
  \texttt{copyedit\_report.tex},
  \texttt{review\_report.tex},
  \texttt{review\_verdict.json},
  opt.\ \texttt{claim\_traceability.json}
& Manuscript governance and revision history for stricter runs
& \texttt{--enforce-editorial-\allowbreak artifacts} \\
\hline
Operational bookkeeping &
  \texttt{checkpoints.db},
  \texttt{run\_token\_usage.json},
  \texttt{budget\_state.json},
  \texttt{budget\_ledger.jsonl},
  \texttt{inter\_agent\_messages/}
& Resume, budget tracking, post-hoc inspection
& Operational trace \\
\hline
Mode-specific artifacts &
  \texttt{experiment\_workspace/*},
  \texttt{experiment\_runs/*},
  \texttt{math\_workspace/*},
  \texttt{counsel\_sandboxes/*},
  \texttt{tree\_search\_state.json},
  \texttt{tree\_branches/*}
& Evidence from empirical, mathematical, counsel, and tree-search modes
& Checked by downstream stages \\
\hline
\end{tabular}
\caption{Artifact bundles in the current \textbf{pAI}\texttt{/MSc} release. The first two rows define the minimal paper-facing contract; remaining rows extend it by run mode.}
\label{tab:artifact_contract}
\end{table}

In stricter paper-oriented modes, validation does more than check file existence. The final gate can also enforce internal review thresholds, paper-quality checks, and, when theorem-oriented or editorial modes are active, additional consistency requirements on theorem dependencies, acceptance status, and claim traceability. These checks are useful because they prevent the system from calling an obviously incomplete run a success. They should not, however, be confused with guarantees of scientific correctness.

A further design choice is that proofreading and reviewer stages emit editable source artifacts such as \texttt{copyedit\_report.tex} and \texttt{review\_report.tex}. Validation is keyed to those source artifacts rather than only to rendered PDFs. This keeps the workflow aligned with a manuscript-development process rather than a screenshot-like end product.

In summary, the artifact contract serves three purposes. It makes the pipeline legible to a human researcher, it provides a stable external interface for resume and post-hoc debugging, and it lets this report state a narrow but defensible notion of success: \textbf{pAI}\texttt{/MSc} can reliably produce and validate a structured research workspace, even though the scientific content of that workspace still requires human scrutiny.

These observations shaped every architectural decision described in the following sections.


\section{Mechanisms for Quality and Rigor}
\label{sec:quality}

Every multi-agent system has stages. What distinguishes \textbf{pAI}\texttt{/MSc} is the \emph{care} baked into pushing quality at each stage. This section describes six concrete mechanisms---each a direct response to the engineering lessons in Section~\ref{sec:requirements}---that together bias the system toward rigorous establishment rather than fluent generation.

\subsection{Persona council: debate with purpose}
\label{sec:persona}

Single-agent ideation tends to converge too quickly to generic or aesthetically plausible plans (Lesson~1 in Section~\ref{sec:requirements}). Our response is the \texttt{persona\_council}: a structured debate among three agents with distinct, competing goals.

\begin{itemize}[itemsep=0.15em,topsep=0.15em,parsep=0pt]
    \item \textbf{Practical Compass.} Pushes for timely, practitioner-relevant research questions. Its job is to keep the project grounded in problems that real researchers care about.
    \item \textbf{Rigor \& Novelty.} Pushes for rigorous, potentially novel mathematical theory. Its job is to ensure that the formal content is strong enough to survive serious scrutiny.
    \item \textbf{Narrative Architect.} Pushes for the strongest narrative arc and the right alignment---or deliberate disalignment---with existing folklore. Its job is to make the paper tell a story that is both accurate and compelling.
\end{itemize}

The three personas do not simply vote. They debate through structured rounds, and synthesis rules force explicit conflict resolution: disagreements must be surfaced and adjudicated, not papered over by averaging. The result is not consensus for its own sake, but a research plan whose tradeoffs have been examined from multiple angles.

This mechanism directly addresses Lesson~2 from Section~\ref{sec:requirements}: giving each agent an explicit, competing objective consistently produced better decompositions and more informative disagreements than a single planner trying to ideate, justify, and plan everything at once.

\subsection{Adversarial novelty falsification}
\label{sec:novelty}

The literature review agent's default stance is \emph{skepticism}---it assumes the central claim is already known and tries to find the evidence. This is not ``find related work.'' It is ``try to kill the novelty claim.''

The agent assigns each claim one of four statuses:
\begin{center}
\texttt{OPEN} \quad\textbar\quad \texttt{PARTIAL} \quad\textbar\quad \texttt{KNOWN} \quad\textbar\quad \texttt{EQUIVALENT\_KNOWN}
\end{center}
Multi-source search---including deep-research APIs, semantic scholar queries, and citation traversal---is employed not to decorate the introduction with related work, but to build a genuine case for or against novelty. A claim rated \texttt{KNOWN} or \texttt{EQUIVALENT\_KNOWN} triggers a rethink: the system must either reformulate the contribution or provide explicit evidence that the existing result does not subsume the new one.

This mechanism matters because one of the most common failure modes of automated research is confident production of work that is, in fact, already established. By making skepticism the default, the system forces early confrontation with the prior literature rather than late-stage embarrassment.

\subsection{Theory--experiment independence}
\label{sec:independence}

A natural design instinct is to maximize context sharing between the mathematical and experimental tracks. In our experience, this improves local alignment but weakens both tracks (Lesson~3 in Section~\ref{sec:requirements}).

The non-obvious design: share \emph{less} context, get \emph{better} papers. The theory and experiment tracks share an explicit decomposition (\texttt{track\_decomposition.json}) and common contribution claims, but they do \emph{not} continuously share unrestricted working context. Instead, they exchange information at controlled synchronization points---especially before synthesis and at the beginning of a new cycle.

Why does this work? When theory and experiments share too much context too early, they tend to collapse toward the same framing, inherit each other's blind spots, and produce intermediate outputs that are better matched but individually less strong. Independence preserves the possibility that the experiment track discovers something the theory track missed, or vice versa. The merge and duality stages then force reconciliation, but only after each track has had the chance to develop its own strongest argument.

\subsection{Reviewer hard blockers}
\label{sec:hardblockers}

The internal reviewer agent does not produce a single smooth score. It evaluates five binary ``hard blockers'' (B1--B5):

\begin{enumerate}[label=\textbf{B\arabic*},itemsep=0.15em]
    \item No explicit research questions stated in the introduction.
    \item No evidence pointers for key takeaways (claims without traceable support).
    \item Placeholders, stubs, or unfinished sections remain in the manuscript.
    \item Critical experimental results are missing or unreported.
    \item Formal claims lack proof sketches or verification traces.
\end{enumerate}

If \emph{any} hard blocker is true, the reviewer score is capped at~4 regardless of prose quality. A score of~8 or above is assigned only when the manuscript is genuinely approaching publication readiness. This shows that the system does not rubber-stamp its own output: a fluent but incomplete paper is flagged as incomplete, and the workflow routes back to revision.

Hard blockers are the mechanism through which the design requirement of separating structural validation from scientific truth (Section~\ref{sec:requirements}) becomes operationally concrete in the editorial phase.

\subsection{Multi-model counsel as structured disagreement}
\label{sec:counsel_quality}

Counsel is not redundancy for robustness. It is structured disagreement for quality.

When counsel is enabled, a specialist stage proceeds as follows. Three frontier models---in the default configuration, Claude Opus 4.6, GPT-5.4, and Gemini 3 Pro Preview---each work in an isolated sandbox copy of the workspace. They produce independent candidate outputs without seeing each other's work. The candidates are then exposed to multiple debate rounds in which each model critiques the others. Finally, a synthesis model (by default Claude Sonnet 4.6) integrates the strongest elements into a single authoritative output that is promoted back to the main workspace.

The goal is to surface alternative reasoning, not to vote. A majority-rules scheme would converge to the blandest common denominator. Instead, the synthesis model must explicitly address disagreements, choosing between alternatives with stated rationale. Engineering discipline is maintained through circuit breakers and quorum logic: if models fail, time out, or produce degenerate outputs, the system falls back gracefully rather than blocking the entire pipeline.

This design targets stages where disagreement is informative---literature review, experiment design, and manuscript writing---and it changes the failure mode from ``one model silently committed to one line of reasoning'' to ``multiple models exposed and critiqued alternative drafts before promotion.''

\subsection{Tree search over proof strategies}
\label{sec:treesearch}

Proof construction has genuine combinatorial structure, and a single linear pass through a proof attempt is often insufficient. When tree search is enabled, the theory track's prover stage is replaced by a DAG-layered best-first search controller.

The controller selects frontier claims from the claim graph, generates alternative proof strategies for each, and scores the resulting branches using a composite function:

\begin{center}
\begin{tabular}{rl}
40\% & LLM-estimated promise of the strategy \\
25\% & Impact on the claim graph (how many downstream claims are unblocked) \\
15\% & Cost efficiency (expected tokens per unit of progress) \\
10\% & Depth penalty (prefer shallower proofs) \\
10\% & Sibling diversity (penalize strategies too similar to existing branches)
\end{tabular}
\end{center}

The top-scoring candidates are executed in forked workspaces. Failed branches can spawn debugging children up to a configurable depth, while low-scoring branches are pruned. The corresponding artifacts---\texttt{tree\_search\_state.json} and \texttt{tree\_branches/}---make the entire search process inspectable rather than hidden inside one monolithic prover call.

The scoring formula itself reflects the care behind the system: it is not a generic ``try again'' loop, but a deliberate balance between exploration breadth, graph-level impact, and computational discipline.

\section{The System}
\label{sec:system}

\begin{figure}[p]
\centering
\resizebox{!}{0.95\textheight}{%
\begin{tikzpicture}[
    node distance=0.35cm and 0.6cm,
    every node/.style={font=\footnotesize},
    agent/.style={rectangle, draw, rounded corners=2pt, minimum width=2.0cm, minimum height=0.55cm, fill=blue!6, align=center},
    counsel/.style={agent, draw=violet!70, densely dashed, thick},
    gate/.style={diamond, draw, minimum width=1.2cm, minimum height=0.7cm, fill=orange!10, align=center, inner sep=1pt, font=\footnotesize},
    track/.style={rectangle, draw=gray!60, rounded corners=5pt, inner sep=6pt, fill=gray!3},
    endnode/.style={circle, draw, fill=green!15, minimum size=0.6cm, font=\footnotesize\bfseries},
    arr/.style={->, >=stealth, semithick},
    loop/.style={->, >=stealth, semithick, red!60!black, densely dashed},
    lbl/.style={font=\tiny, fill=white, inner sep=1pt},
    phaselbl/.style={font=\scriptsize\bfseries\sffamily, text=gray!70!black},
]

\node[phaselbl] (ph1) {Phase 1};
\node[counsel, right=0.4cm of ph1] (persona) {Persona Council\\[-1pt]{\tiny(3-persona debate)}};
\node[counsel, below=of persona] (litrev) {Literature Review};
\node[gate, below=of litrev] (litgate) {Lit\\Gate};

\node[counsel, below=0.3cm of litgate] (brainstorm) {Brainstorm};
\node[counsel, below=of brainstorm] (goals) {Formalize Goals};
\node[gate, below=of goals] (trackrouter) {Track\\Router};

\node[counsel, below left=0.7cm and 2.0cm of trackrouter] (mathlit) {Math Lit};
\node[counsel, below=of mathlit] (mathprop) {Proposer};
\node[counsel, below=of mathprop] (mathprover) {Prover\\[-1pt]{\tiny(or TreeSearch)}};
\node[counsel, below=of mathprover] (mathrig) {Rigorous Verif.};
\node[counsel, below=of mathrig] (mathemp) {Empirical Verif.};
\node[counsel, below=of mathemp] (prooftrans) {Proof Transcr.};

\begin{scope}[on background layer]
\node[track, fit=(mathlit)(prooftrans), label={[font=\scriptsize\itshape]above:Theory Track}] (theorybox) {};
\end{scope}

\node[counsel, below right=0.7cm and 2.0cm of trackrouter] (explit) {Exp Literature};
\node[counsel, below=of explit] (expdesign) {Exp Design};
\node[counsel, below=of expdesign] (expexec) {Experimentation};
\node[counsel, below=of expexec] (expverify) {Exp Verification};
\node[counsel, below=of expverify] (exptrans) {Exp Transcr.};

\begin{scope}[on background layer]
\node[track, fit=(explit)(exptrans), label={[font=\scriptsize\itshape]above:Experiment Track}] (expbox) {};
\end{scope}

\node[agent, below=1.2cm of prooftrans, xshift=2.0cm] (merge) {Track Merge};
\node[gate, below=of merge] (verifycomp) {Verify\\Compl.};

\node[counsel, below=0.3cm of verifycomp] (formalresults) {Formalize Results};
\node[agent, below=of formalresults] (duality) {Duality Check};
\node[gate, below=of duality] (dualitygate) {Duality\\Gate};

\node[counsel, below=0.3cm of dualitygate] (resourceprep) {Resource Prep};
\node[counsel, below=of resourceprep] (writeup) {Writeup};
\node[counsel, below=of writeup] (proofread) {Proofreading};
\node[counsel, below=of proofread] (reviewer) {Reviewer};
\node[gate, below=of reviewer] (valgate) {Valid.\\Gate};
\node[endnode, below=0.3cm of valgate] (endn) {END};

\node[agent, right=1.5cm of dualitygate] (followuplit) {Follow-up\\Lit Review};

\draw[arr] (persona) -- (litrev);
\draw[arr] (litrev) -- (litgate);
\draw[arr] (litgate) -- node[lbl,right] {feasible} (brainstorm);
\draw[arr] (brainstorm) -- (goals);
\draw[arr] (goals) -- (trackrouter);
\draw[arr] (trackrouter) -| node[lbl, pos=0.25, above] {\scriptsize theory} (mathlit);
\draw[arr] (trackrouter) -| node[lbl, pos=0.25, above] {\scriptsize empirical} (explit);
\draw[arr] (mathlit) -- (mathprop);
\draw[arr] (mathprop) -- (mathprover);
\draw[arr] (mathprover) -- (mathrig);
\draw[arr] (mathrig) -- (mathemp);
\draw[arr] (mathemp) -- (prooftrans);
\draw[arr] (explit) -- (expdesign);
\draw[arr] (expdesign) -- (expexec);
\draw[arr] (expexec) -- (expverify);
\draw[arr] (expverify) -- (exptrans);
\draw[arr] (prooftrans) |- (merge);
\draw[arr] (exptrans) |- (merge);
\draw[arr] (merge) -- (verifycomp);
\draw[arr] (verifycomp) -- node[lbl,right] {\scriptsize complete} (formalresults);
\draw[arr] (formalresults) -- (duality);
\draw[arr] (duality) -- (dualitygate);
\draw[arr] (dualitygate) -- node[lbl,right] {\scriptsize pass} (resourceprep);
\draw[arr] (resourceprep) -- (writeup);
\draw[arr] (writeup) -- (proofread);
\draw[arr] (proofread) -- (reviewer);
\draw[arr] (reviewer) -- (valgate);
\draw[arr] (valgate) -- node[lbl,right] {\scriptsize pass} (endn);

\draw[loop] (litgate.west) -- +(-0.9,0) |- node[lbl, pos=0.25, left] {\scriptsize infeasible} (persona.west);
\draw[loop] (verifycomp.west) -- +(-3.8,0) |- node[lbl, pos=0.12, left] {\scriptsize incomplete} (goals.west);
\draw[loop] (verifycomp.east) -- +(3.8,0) |- node[lbl, pos=0.12, right] {\scriptsize rethink} (brainstorm.east);
\draw[loop] (dualitygate) -- node[lbl, above] {\scriptsize fail} (followuplit);
\draw[loop] (followuplit) |- (brainstorm.east);
\draw[loop] (valgate.west) -- +(-1.2,0) |- node[lbl, pos=0.25, left] {\scriptsize fail} (writeup.west);

\node[below right=0.4cm and -0.5cm of endn, font=\tiny, anchor=north west] (leg1) {%
  \tikz[baseline=-0.5ex]{\node[counsel, minimum width=0.8cm, minimum height=0.3cm] {};}\; Counsel-eligible \quad
  \tikz[baseline=-0.5ex]{\node[gate, minimum width=0.5cm, minimum height=0.3cm] {};}\; Gate \quad
  \tikz[baseline=-0.5ex]{\draw[loop] (0,0) -- (0.6,0);}\; Loopback%
};

\end{tikzpicture}
}
\caption{The \textbf{pAI}\texttt{/MSc} execution graph. Dashed violet borders mark counsel-eligible agents. Red dashed arrows are loopbacks triggered by gate failures. Theory and experiment tracks run in parallel when both are selected.}
\label{fig:overview}
\end{figure}
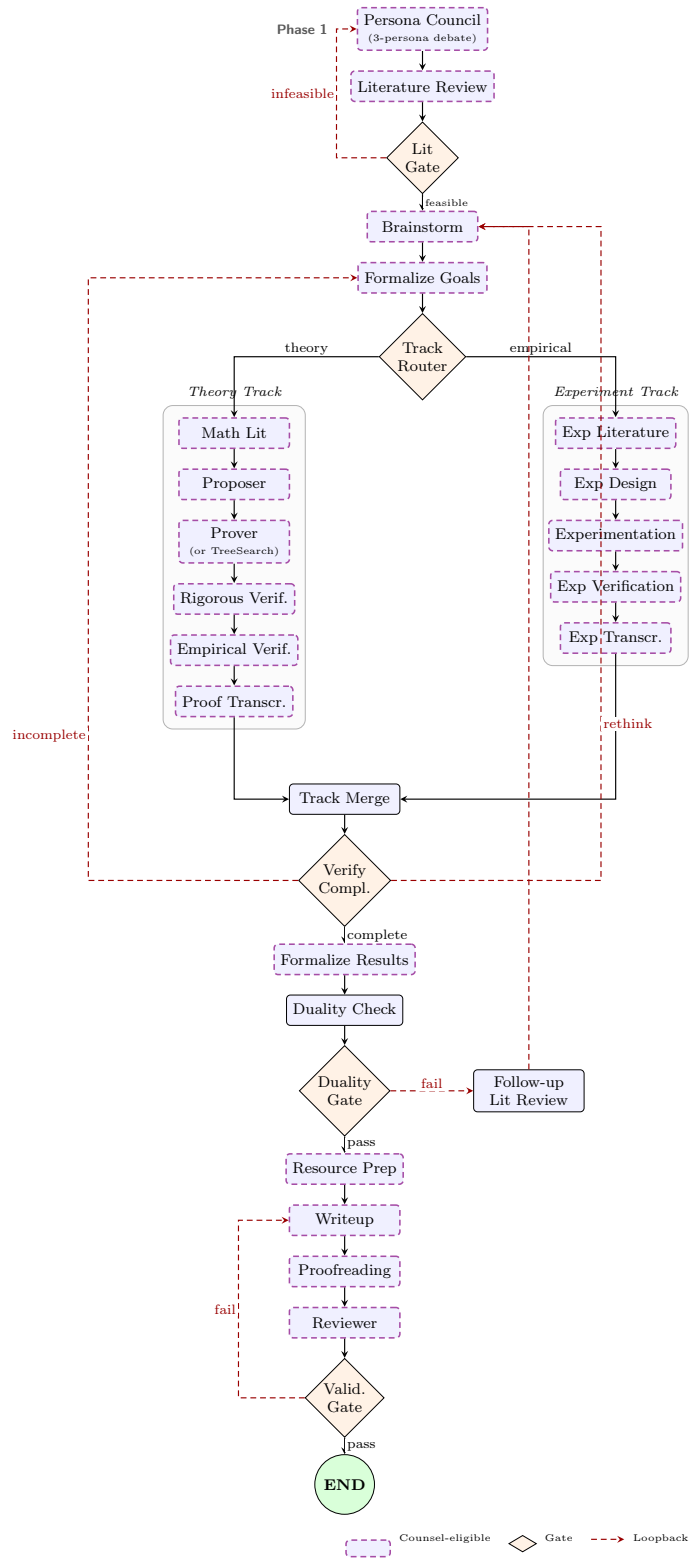

This section gives a single, self-contained walkthrough of \textbf{pAI}\texttt{/MSc}. The workflow contains 23 specialist agents and 30 total graph nodes; the additional nodes are routing gates and control points shown in Figure~\ref{fig:overview}. We describe the pipeline phases, reference the artifact contract from Section~\ref{sec:requirements}, summarize the optional modules, and close with the operational model. Readers who want full implementation details should consult Appendix~\ref{sec:architecture_reference}.

\subsection{Pipeline phases}
\label{sec:pipeline}

At the core of the release is a fixed LangGraph workflow. ``Fixed'' refers to the execution topology: the graph always traverses the same high-level stages and routing logic, even though the generated content remains stochastic. The system is not deterministic in output, but it is explicit and repeatable in control flow. The graph is organized into six phases.

\paragraph{Phase 1: Discovery and feasibility screening.}
The run begins with a \texttt{persona\_council} stage (Section~\ref{sec:persona}), followed by \texttt{literature\_review\_agent}, a feasibility gate, \texttt{brainstorm\_agent}, and \texttt{formalize\_goals\_agent}. This phase grounds the task in prior work, checks whether the direction appears feasible, and converts a high-level objective into a concrete technical plan.

\paragraph{Phase 2: Track planning and routing.}
Planning writes a decomposition into theory, experiment, or mixed tracks, recorded in \texttt{paper\_workspace/\allowbreak track\_decomposition.json}. The track router uses this decomposition to decide which execution branches to launch.

\paragraph{Phase 3: Parallel technical execution.}
When theory work is selected and theorem-oriented modules are enabled, the graph runs the theory path: mathematical literature review, proposal, proving, rigorous verification, empirical verification, and proof transcription. When empirical work is selected, it runs experiment literature review, design, execution, verification, and transcription. The two tracks can run independently or in parallel.

\paragraph{Phase 4: Completion verification and results formalization.}
Track outputs are merged and passed to \texttt{verify\_completion}, which routes the run in three directions: proceed, iterate because the work is incomplete, or rethink the direction more fundamentally. On success, \texttt{formalize\_results\_agent} produces a structured assessment of what has been established.

\paragraph{Phase 5: Internal consistency and follow-up.}
A duality check can be applied after results formalization. Failure triggers follow-up literature work and renewed planning; success moves the run to paper production.

\paragraph{Phase 6: Paper production and editorial quality assurance.}
The pipeline prepares resources, writes the paper, proofreads it, runs an internal reviewer, and applies a final validation gate. Validation failure routes back to writeup for revision before a success state is reported.

\subsection{Artifact contract}
\label{sec:artifact_contract_ref}

All six phases communicate through the artifact contract described in Section~\ref{sec:requirements} and summarized in Table~\ref{tab:artifact_contract}. A run is not treated as successful merely because it produces fluent prose; it must materialize the required named artifacts for discovery, planning, execution, synthesis, and editorial review. In stricter modes, the final gate additionally enforces internal review thresholds, theorem-dependency consistency, and claim traceability.

\subsection{Optional modules}
\label{sec:optional_modules}

\paragraph{Counsel.}
When \texttt{--enable-counsel} is active, selected specialist stages become a multi-model protocol with sandboxing, debate rounds, synthesis, and artifact promotion. Three frontier models work in isolated sandboxes, critique each other's outputs, and a synthesis model promotes a single authoritative result. Details are in Section~\ref{sec:counsel_quality}.

\paragraph{Tree search.}
When \texttt{--enable-tree-search} is active, the linear proof stage in the theory track is replaced by a DAG-layered best-first controller that explores multiple proof strategies in parallel. See Section~\ref{sec:treesearch}.

\paragraph{Mathematical reasoning pipeline.}
When \texttt{--enable-math-agents} is active, the theory track gains a full claim graph, proof workspace, lemma library, and staged verification. The math workspace stores claims, proofs, and verification traces as structured artifacts rather than free-form text.

\paragraph{Campaign orchestration.}
The optional campaign layer uses heartbeat scripts, cron, or OpenClaw to repeatedly launch or resume runs, validate stage artifacts, distill stage memory, track budget, attempt autonomous repair, and notify the user. This extends the artifact-centric philosophy beyond a single run.

\subsection{Operational model}
\label{sec:operational_model}

\textbf{pAI}\texttt{/MSc} is designed as a human-on-the-loop system. A \emph{steer} is a deliberate human intervention---distinct from the graph's own automatic loopbacks---that changes the direction, constraints, or scope of a run. The current release exposes steering through: the initial task specification; a TCP control socket and HTTP API during execution; resume from a named stage; injection of context files through an \texttt{inputs/} directory; and campaign-level control for multi-stage pipelines.

Every completed stage is persisted to SQLite in \texttt{checkpoints.db}, making long runs restartable and allowing downstream tools to reason about partially completed workspaces. Budget tracking is implemented through \texttt{budget\_state.json}, \texttt{budget\_ledger.jsonl}, and \texttt{run\_token\_usage.json}.

\begin{table}[t]
\centering
\small
\setlength{\tabcolsep}{4pt}
\begin{tabular}{|L{3.8cm}|L{1.8cm}|L{2.0cm}|L{3.6cm}|}
\hline
\textbf{Configuration} & \textbf{Typical cost} & \textbf{Typical runtime} & \textbf{Notes} \\
\hline
Quickstart (markdown, no counsel) & \$2--10 & 15--40 min & Single-model run, light paper draft \\
\hline
Base pipeline + \LaTeX{}/PDF & \$10--40 & 30--90 min & Requires \LaTeX{} toolchain \\
\hline
Base pipeline + math agents & \$20--60 & 60--150 min & Adds theorem-oriented stages \\
\hline
Counsel mode & \$50--200 & 2--5 hrs & Multi-model debate and synthesis \\
\hline
Tree search (no counsel) & \$60--180 & 2--4 hrs & Parallel proof exploration \\
\hline
Tree search + counsel & \$200--600 & 4--10 hrs & Highest-cost single-run mode \\
\hline
Full paper campaign & \$100--400 & 6--12 hrs & Multi-stage orchestration over several runs \\
\hline
\end{tabular}
\caption{Runtime and spend envelopes for representative \textbf{pAI}\texttt{/MSc} configurations. Actual values vary with task scope, model choice, revision loops, and experiment intensity.}
\label{tab:runtime_cost}
\end{table}

\FloatBarrier  
\section{Related Work}
\label{sec:related_work}

The literature relevant to \textbf{pAI}\texttt{/MSc} spans four lines. We position the system briefly here; a comprehensive survey is in Appendix~\ref{sec:extended_related_work}.

\paragraph{Scholarly writing assistants.}
Early LLM-era work focused on citation-grounded text generation rather than full research automation. Systems such as CiteBench \citep{funkquist2023citebench}, ChatCite \citep{li2024chatcite}, ScholarCopilot \citep{wang2025scholarcopilot}, SurveyGen \citep{bao2025surveygen}, and editor-native tools like OverleafCopilot \citep{wen2024overleafcopilot} and PaperDebugger \citep{hou2025paperdebugger} established that retrieval quality and claim--evidence linkage matter more than surface fluency. These are important precursors, but their primary artifact is a section or draft rather than a coordinated research program.

\paragraph{Manuscript-first research-to-paper agents.}
The closest predecessors are end-to-end systems targeting research-to-paper execution. Data-to-paper \citep{ifargan2025datatopaper} emphasized provenance and backward traceability. The AI Scientist \citep{lu2024aiscientist} and its successor \citep{yamada2025aiscientistv2} pushed toward fully autonomous ML research with tree search and automated review. Agent Laboratory \citep{schmidgall2025agentlaboratory} showed that human feedback at stage boundaries materially improves quality. CycleResearcher \citep{weng2025cycleresearcher} coupled generation with review loops, AI-Researcher \citep{tang2025airesearcher} introduced Scientist-Bench for open-ended evaluation, and AgentRxiv \citep{schmidgall2025agentrxiv} enabled cross-run collaboration through a preprint-server abstraction. freephdlabor \citep{li2025freephdlabor} argued for fully dynamic, user-customizable workflows. Relative to these systems, \textbf{pAI}\texttt{/MSc} places more weight on a tighter artifact contract, theory--experiment coordination, and bounded workflow control.

\paragraph{Autonomous discovery and open platforms.}
Broader systems target scientific discovery rather than manuscript production. AI co-scientist \citep{gottweis2025aicoscientist} debates and refines hypotheses in biomedicine; InternAgent \citep{internagentteam2025internagent,feng2026internagent15} pursues long-horizon discovery across domains; AlphaEvolve \citep{novikov2025alphaevolve} and Mathematical Exploration at Scale \citep{georgiev2025mathematicalexploration} target algorithmic and mathematical discovery; and DeepInnovator \citep{fan2026deepinnovator} trains models for innovative idea generation. On the platform side, ClawdLab and Beach.Science \citep{weidener2026clawdlab} introduce governance, auditability, and decentralized coordination for scientific-agent ecosystems. These systems enlarge the meaning of research automation, but many optimize for open-ended discovery or ecosystem flexibility rather than a governed paper-facing workflow.

\paragraph{Positioning.}
\textbf{pAI}\texttt{/MSc} is closest in spirit to manuscript-first, end-to-end systems, but differs in emphasis. Relative to writing assistants, it covers more of the research lifecycle. Relative to autonomous discovery systems, it places more weight on explicit intermediate artifacts, theory--experiment coordination, and validation gates tied to manuscript production. Relative to dynamic or decentralized platforms, it favors a tightly governed research-to-paper path designed for human-steerable execution in machine learning theory and adjacent quantitative workflows. It should be read not as a general theory of autonomous science, but as an artifact-driven technical contribution to the research-to-paper segment of that broader landscape.

\section{Limitations and Honest Assessment}
\label{sec:limitations}

\subsection{What the system guarantees and what it does not}

The most important distinction in this report is between \emph{structural} and \emph{scientific} success. The system's guarantees are structural and operational; its non-guarantees are scientific and epistemic. Table~\ref{tab:guarantees} makes this boundary explicit.

\begin{table}[t]
\centering
\small
\setlength{\tabcolsep}{4pt}
\begin{tabular}{|L{6.2cm}|L{6.8cm}|}
\hline
\textbf{The system can guarantee} & \textbf{The system cannot guarantee} \\
\hline
Deterministic execution topology: same stages, same routing logic, same loopback structure & Deterministic outputs: two runs with the same task may diverge in content \\
\hline
Artifact-complete workspace: all required files produced and structurally valid & Scientifically correct workspace: claims may contain errors, gaps, or undetected prior art \\
\hline
Explicit validation gates: hard blockers, review thresholds, editorial checks & Calibrated quality scores: internal reviewer scores are routing heuristics, not acceptance-probability estimates \\
\hline
Budget tracking and cost visibility via ledger files & Exact cost prediction: actual provider bills may differ from estimates \\
\hline
Resumable, checkpointed execution with stage-level restart & Idempotent restart: resumed runs may produce different outputs from fresh runs \\
\hline
Human steering surfaces at multiple points in the pipeline & Reduced need for human scientific judgment: novelty, correctness, and citation faithfulness remain human obligations \\
\hline
\end{tabular}
\caption{Guarantees and non-guarantees of the current \textbf{pAI}\texttt{/MSc} release. The central distinction is between structural/operational properties (left) and scientific properties (right).}
\label{tab:guarantees}
\end{table}

\subsection{What is surprisingly automatable}

One of the most interesting lessons is that \emph{research structuring} is more automatable than one might expect. The system can turn a high-level task into a staged workspace with literature synthesis, a research plan, a theory--experiment decomposition, follow-up decisions, and a draft manuscript. Even when the final science requires human scrutiny, this intermediate structuring is already valuable.

A second surprisingly automatable layer is \emph{paper-adjacent discipline}: proofreading, reviewer-style critique, copyediting traces, claim-traceability files, resource inventories, and revision logs. These are tedious but structurally well suited to an artifact-centric pipeline. Some of the ``boring graduate student work'' is indeed becoming tractable---not because the machine understands the science end to end, but because many parts of scientific production are really state-management and transformation problems.

\subsection{What remains stubbornly human}

The hardest parts of research remain strongly human-centered. Novelty judgments are open-world judgments. Strong baseline selection depends on field knowledge and taste. Mathematical correctness often hinges on subtle proof obligations that resist shallow verification. Experimental interpretation depends on whether the chosen measurements actually address the scientific question. Even manuscript quality has a human-dependent layer: deciding what the paper is \emph{really about}, what should be emphasized, and what evidence is persuasive to a given community.

This is why the right control model for the present release is \emph{human on the loop}, not \emph{human out of the loop}. The system can propose, decompose, execute, and draft; the human remains responsible for owning the claims.

\subsection{Toward stronger evaluation}

The next milestone is not ``more features'' but a stronger evaluation program. A convincing benchmark would need repeated runs across multiple task families, explicit logging of human steering interventions, external citation-faithfulness audits, reruns of generated experiments, and expert assessments of theorem and manuscript quality. It should also include ablations isolating the value of counsel, tree search, strict editorial enforcement, and campaign-style repair.

The present release is best understood as an infrastructure milestone. It makes those evaluations possible by standardizing workflow state, artifact contracts, and operating modes. The stronger scientific claims should come only after that measurement program is in place.

\section{Conclusion}
\label{sec:conclusion}

\textbf{pAI}\texttt{/MSc} is best described, in its present form, as a human-on-the-loop, artifact-centric research-to-paper pipeline. Its contribution is not that it solves scientific truth or automates novelty judgment, but that it turns a research task into a traceable workspace with planning artifacts, theory and experiment branches, revision loops, manuscript outputs, and explicit validation gates.

That contribution is already meaningful. It moves the problem from ``how do I keep a long sequence of model calls coherent?'' to ``how do I inspect, guide, and validate a structured research workspace?'' The guarantees, limitations, runtime modes, safety notes, and submission checklist all point to the same conclusion: the present system is a substantial step toward reducing human steering burden, but its outputs still require expert verification before any scientific claim is trusted or submitted.

The next phase should therefore focus on benchmarked evaluation. If future work can pair the current architecture with rigorous measurements of citation faithfulness, experiment reproducibility, theorem reliability, and human steering reduction, then \textbf{pAI}\texttt{/MSc} will provide not only a useful release but also clearer scientific understanding of what parts of research automation are truly becoming tractable.

\newpage

\bibliography{reference_PoggioAI}

\begin{thebibliography}{89}
\providecommand{\natexlab}[1]{#1}
\providecommand{\url}[1]{\texttt{#1}}
\expandafter\ifx\csname urlstyle\endcsname\relax
  \providecommand{\doi}[1]{doi: #1}\else
  \providecommand{\doi}{doi: \begingroup \urlstyle{rm}\Url}\fi

\bibitem[Romera-Paredes et~al.(2024)Romera-Paredes, Barekatain, Novikov, Balog, Kumar, Dupont, Ruiz, Ellenberg, Wang, Fawzi, Kohli, and Fawzi]{funsearchNature2024}
Bernardino Romera-Paredes, Mohammadamin Barekatain, Alexander Novikov, Matej Balog, M.~Pawan Kumar, Emilien Dupont, Francisco J.~R. Ruiz, Jordan~S. Ellenberg, Pengming Wang, Omar Fawzi, Pushmeet Kohli, and Alhussein Fawzi.
\newblock Mathematical discoveries from program search with large language models.
\newblock \emph{Nature}, 625\penalty0 (7995):\penalty0 468--475, 2024.
\newblock \doi{10.1038/s41586-023-06924-6}.
\newblock URL \url{https://www.nature.com/articles/s41586-023-06924-6}.

\bibitem[{Google DeepMind}(2023)]{funsearchRepo}
{Google DeepMind}.
\newblock Funsearch.
\newblock GitHub repository, 2023.
\newblock URL \url{https://github.com/google-deepmind/funsearch}.
\newblock Repository accompanying the FunSearch Nature paper; accessed 2026-03-21.

\bibitem[Novikov et~al.(2025{\natexlab{a}})Novikov, V{\~u}, Eisenberger, Dupont, Huang, Wagner, Shirobokov, Kozlovskii, Ruiz, Mehrabian, Kumar, See, Chaudhuri, Holland, Davies, Nowozin, Kohli, and Balog]{alphaevolve2025}
Alexander Novikov, Ng{\^a}n V{\~u}, Marvin Eisenberger, Emilien Dupont, Po-Sen Huang, Adam~Zsolt Wagner, Sergey Shirobokov, Borislav Kozlovskii, Francisco J.~R. Ruiz, Abbas Mehrabian, M.~Pawan Kumar, Abigail See, Swarat Chaudhuri, George Holland, Alex Davies, Sebastian Nowozin, Pushmeet Kohli, and Matej Balog.
\newblock Alphaevolve: A coding agent for scientific and algorithmic discovery.
\newblock \emph{arXiv preprint arXiv:2506.13131}, 2025{\natexlab{a}}.
\newblock \doi{10.48550/arXiv.2506.13131}.
\newblock URL \url{https://arxiv.org/abs/2506.13131}.

\bibitem[Georgiev et~al.(2025{\natexlab{a}})Georgiev, G{\'o}mez-Serrano, Tao, and Wagner]{mathexplore2025}
Bogdan Georgiev, Javier G{\'o}mez-Serrano, Terence Tao, and Adam~Zsolt Wagner.
\newblock Mathematical exploration and discovery at scale.
\newblock \emph{arXiv preprint arXiv:2511.02864}, 2025{\natexlab{a}}.
\newblock \doi{10.48550/arXiv.2511.02864}.
\newblock URL \url{https://arxiv.org/abs/2511.02864}.

\bibitem[{Google DeepMind}(2025)]{alphaevolveProblemRepo}
{Google DeepMind}.
\newblock Mathematical problem repository for alphaevolve.
\newblock GitHub repository, 2025.
\newblock URL \url{https://github.com/google-deepmind/alphaevolve_repository_of_problems}.
\newblock Repository accompanying the Mathematical exploration and discovery at scale preprint; accessed 2026-03-21.

\bibitem[Nagda et~al.(2026)Nagda, Raghavan, and Thakurta]{ramseyAlphaEvolve2026}
Ansh Nagda, Prabhakar Raghavan, and Abhradeep Thakurta.
\newblock Reinforced generation of combinatorial structures: Ramsey numbers.
\newblock \emph{arXiv preprint arXiv:2603.09172}, 2026.
\newblock \doi{10.48550/arXiv.2603.09172}.
\newblock URL \url{https://arxiv.org/abs/2603.09172}.

\bibitem[Knuth(2026)]{claudesCycles2026}
Donald~E. Knuth.
\newblock Claude's cycles.
\newblock Informal note / PDF on Knuth's preprints page, February 2026.
\newblock URL \url{https://cs.stanford.edu/~knuth/papers/claude-cycles.pdf}.
\newblock Dated 2026-02-28; revised 2026-03-16.

\bibitem[Tao(2025)]{taoErdos1026_2025}
Terence Tao.
\newblock The story of erd{\H{o}}s problem \#1026.
\newblock Blog post on {\it What's New}, December 2025.
\newblock URL \url{https://terrytao.wordpress.com/2025/12/08/the-story-of-erdos-problem-126/}.
\newblock Published 2025-12-08.

\bibitem[Abouzaid et~al.(2026)Abouzaid, Blumberg, Hairer, Kileel, Kolda, Nelson, Spielman, Srivastava, Ward, Weinberger, and Williams]{firstProof2026}
Mohammed Abouzaid, Andrew~J. Blumberg, Martin Hairer, Joe Kileel, Tamara~G. Kolda, Paul~D. Nelson, Daniel Spielman, Nikhil Srivastava, Rachel Ward, Shmuel Weinberger, and Lauren Williams.
\newblock First proof.
\newblock \emph{arXiv preprint arXiv:2602.05192}, 2026.
\newblock \doi{10.48550/arXiv.2602.05192}.
\newblock URL \url{https://arxiv.org/abs/2602.05192}.

\bibitem[{First Proof Project}(2026)]{firstBatchSite2026}
{First Proof Project}.
\newblock First batch.
\newblock Project website, February 2026.
\newblock URL \url{https://1stproof.org/first-batch.html}.
\newblock First-batch page; site lists February 2026 release context; accessed 2026-03-21.

\bibitem[{OpenAI}(2026)]{openaiFirstProofSubmissions2026}
{OpenAI}.
\newblock Our first proof submissions.
\newblock OpenAI research page, February 2026.
\newblock URL \url{https://openai.com/index/first-proof-submissions/}.
\newblock Published 2026-02-20.

\bibitem[Zhang and Ma(2026)]{zhangMaFirstProofZenodo2026}
Wenlin Zhang and Haobo Ma.
\newblock Lean 4 formal verification of 8/10 \#1stproof problems: Complete proofs with ai--human pipeline, partial qed for q4 \& q6.
\newblock Zenodo preprint, February 2026.
\newblock URL \url{https://zenodo.org/records/18635744}.
\newblock Created 2026-02-13. Zenodo also lists a second record with the same title and metadata at DOI 10.5281/zenodo.18635110.

\bibitem[{OpenAI}(2025{\natexlab{a}})]{openaiGPT52ScienceMath2025}
{OpenAI}.
\newblock Advancing science and math with gpt-5.2.
\newblock OpenAI publication, December 2025{\natexlab{a}}.
\newblock URL \url{https://openai.com/index/gpt-5-2-for-science-and-math}.
\newblock Published 2025-12-11.

\bibitem[Sellke and Yin(2025)]{sellkeYinLearningCurve2025}
Mark Sellke and Steven Yin.
\newblock On learning-curve monotonicity for maximum likelihood estimators.
\newblock \emph{arXiv preprint arXiv:2512.10220}, 2025.
\newblock \doi{10.48550/arXiv.2512.10220}.
\newblock URL \url{https://arxiv.org/abs/2512.10220}.

\bibitem[{Math, Inc.}(n.d.{\natexlab{a}})]{mathIncGauss}
{Math, Inc.}
\newblock Introducing gauss, an agent for autoformalization.
\newblock Company blog post, n.d.{\natexlab{a}}.
\newblock URL \url{https://www.math.inc/gauss}.
\newblock Undated page; accessed 2026-03-21.

\bibitem[{Math, Inc.}(n.d.{\natexlab{b}})]{mathIncStrongPNTSite}
{Math, Inc.}
\newblock Strong pnt.
\newblock Project page, n.d.{\natexlab{b}}.
\newblock URL \url{https://math-inc.github.io/strongpnt/}.
\newblock Undated page; accessed 2026-03-21.

\bibitem[{Math, Inc.}(n.d.{\natexlab{c}})]{mathIncStrongPNTRepo}
{Math, Inc.}
\newblock strongpnt.
\newblock GitHub repository, n.d.{\natexlab{c}}.
\newblock URL \url{https://github.com/math-inc/strongpnt}.
\newblock Repository for the Strong PNT formalization; accessed 2026-03-21.

\bibitem[Lichtman(2025)]{fieldsGaussPNTTalk2025}
Jared~Duker Lichtman.
\newblock Gauss -- an agentic formalization of the prime number theorem.
\newblock Fields Institute talk page, October 2025.
\newblock URL \url{https://www.fields.utoronto.ca/talks/Gauss-agentic-formalization-Prime-Number-Theorem}.
\newblock Talk date: 2025-10-28.

\bibitem[Sothanaphan(2026)]{sothanaphanErdos728_2026}
Nat Sothanaphan.
\newblock Resolution of erd{\H{o}}s problem \#728: a writeup of aristotle's lean proof.
\newblock \emph{arXiv preprint arXiv:2601.07421}, 2026.
\newblock \doi{10.48550/arXiv.2601.07421}.
\newblock URL \url{https://arxiv.org/abs/2601.07421}.

\bibitem[{Harmonic}(2026)]{harmonicErdos728Post2026}
{Harmonic}.
\newblock Today marks a momentous milestone for ai and mathematics.
\newblock X post, January 2026.
\newblock URL \url{https://x.com/HarmonicMath/status/2008693723413225814}.
\newblock Posted 2026-01-06; dynamic-source metadata should be rechecked before camera-ready copy if cited in the main text.

\bibitem[Bloom(2026{\natexlab{a}})]{bloomErdos728_2026}
Thomas~F. Bloom.
\newblock Erd{\H{o}}s problem \#728.
\newblock ErdosProblems.com entry, January 2026{\natexlab{a}}.
\newblock URL \url{https://www.erdosproblems.com/728}.
\newblock Page last edited 2026-01-06; accessed 2026-03-21.

\bibitem[Bloom(2026{\natexlab{b}})]{bloomErdos729_2026}
Thomas~F. Bloom.
\newblock Erd{\H{o}}s problem \#729.
\newblock ErdosProblems.com entry, January 2026{\natexlab{b}}.
\newblock URL \url{https://www.erdosproblems.com/729}.
\newblock Page last edited 2026-01-11; accessed 2026-03-21.

\bibitem[Bloom(2026{\natexlab{c}})]{bloomErdos397_2026}
Thomas~F. Bloom.
\newblock Erd{\H{o}}s problem \#397.
\newblock ErdosProblems.com entry, January 2026{\natexlab{c}}.
\newblock URL \url{https://www.erdosproblems.com/397}.
\newblock Page last edited 2026-01-12; accessed 2026-03-21.

\bibitem[{teorth}(n.d.)]{erdosProblemDatabaseRepo}
{teorth}.
\newblock Erd{\H{o}}s problem database.
\newblock GitHub repository, n.d.
\newblock URL \url{https://github.com/teorth/erdosproblems}.
\newblock Repository README accessed 2026-03-21.

\bibitem[{GIGAZINE}(2025)]{gigazineOpenAIErdosClaims2025}
{GIGAZINE}.
\newblock An openai researcher posted that ``gpt-5 has solved an unsolved mathematical problem,'' but it turned out that the problem had already been solved, leading to ridicule from rival developers, including google deepmind ceo demis hassabis.
\newblock News article, October 2025.
\newblock URL \url{https://gigazine.net/gsc_news/en/20251020-openai-researcher-announced-gpt-5-math-breakthrough/}.
\newblock Published 2025-10-20.

\bibitem[Vogelaar(2025)]{brightOpenAIErdosClaims2025}
Erwin Vogelaar.
\newblock G{\^e}nant: Openai beweert dat chatgpt wiskundeproblemen oplost, maar dat klopt niet.
\newblock Bright.nl news article, October 2025.
\newblock URL \url{https://www.bright.nl/nieuws/1703437/g-nant-openai-beweert-dat-chatgpt-wiskundeproblemen-oplost-maar-dat-klopt-niet.html}.
\newblock Published 2025-10-20; accessed 2026-03-21.

\bibitem[Schwartz(2026{\natexlab{a}})]{schwartzVibePhysics2026}
Matthew~D. Schwartz.
\newblock Vibe physics: The {AI} grad student.
\newblock Anthropic Science Blog, March 2026{\natexlab{a}}.
\newblock URL \url{https://www.anthropic.com/research/vibe-physics}.
\newblock Accessed: 2026-03-24.

\bibitem[Schwartz(2026{\natexlab{b}})]{schwartzCParameter2026}
Matthew~D. Schwartz.
\newblock Resummation of the c-parameter sudakov shoulder using effective field theory.
\newblock \emph{arXiv preprint arXiv:2601.02484}, 2026{\natexlab{b}}.
\newblock \doi{10.48550/arXiv.2601.02484}.
\newblock URL \url{https://arxiv.org/abs/2601.02484}.

\bibitem[Lipton and Steinhardt(2019)]{liptonSteinhardt2019Troubling}
Zachary~C. Lipton and Jacob Steinhardt.
\newblock Troubling trends in machine learning scholarship.
\newblock \emph{Queue}, 17\penalty0 (1), 2019.
\newblock \doi{10.1145/3317287.3328534}.
\newblock URL \url{https://doi.org/10.1145/3317287.3328534}.
\newblock ACM Queue article; multiple secondary indexes report pages 45--77, but page/article-number formatting varies across services, so pages are omitted here deliberately.

\bibitem[Gelman(2019)]{gelman2019TroublingBlog}
Andrew Gelman.
\newblock ``troubling trends in machine learning scholarship''.
\newblock Statistical Modeling, Causal Inference, and Social Science blog, September 2019.
\newblock URL \url{https://statmodeling.stat.columbia.edu/2019/09/30/troubling-trends-in-machine-learning-scholarship/}.
\newblock Blog commentary pointing to Lipton and Steinhardt and discussing hype, ``provably'' language, and advertisement-like ML papers. Accessed 2026-03-21.

\bibitem[Pineau et~al.(2021)Pineau, Vincent-Lamarre, Sinha, Larivi{\`e}re, Beygelzimer, d'Alch{\'e} Buc, Fox, and Larochelle]{pineauEtAl2021ImprovingReproducibility}
Joelle Pineau, Philippe Vincent-Lamarre, Koustuv Sinha, Vincent Larivi{\`e}re, Alina Beygelzimer, Florence d'Alch{\'e} Buc, Emily Fox, and Hugo Larochelle.
\newblock Improving reproducibility in machine learning research ({A} report from the {N}eur{IPS} 2019 reproducibility program).
\newblock \emph{Journal of Machine Learning Research}, 22\penalty0 (164):\penalty0 1--20, 2021.
\newblock URL \url{https://www.jmlr.org/papers/v22/20-303.html}.
\newblock Title checked directly against the JMLR PDF first page.

\bibitem[Bean et~al.(2025)Bean, Kearns, Romanou, Hafner, Mayne, Batzner, Foroutan, Schmitz, Korgul, Batra, Deb, Beharry, Emde, Foster, Gausen, Grandury, Han, Hofmann, Ibrahim, Kim, Kirk, Lin, Liu, Luettgau, Magomere, Rystr{\o}m, Sotnikova, Yang, Zhao, Bibi, Bosselut, Clark, Cohan, Foerster, Gal, Hale, Raji, Summerfield, Torr, Ududec, Rocher, and Mahdi]{beanEtAl2025MeasuringWhatMatters}
Andrew~M. Bean, Ryan~Othniel Kearns, Angelika Romanou, Franziska~Sofia Hafner, Harry Mayne, Jan Batzner, Negar Foroutan, Chris Schmitz, Karolina Korgul, Hunar Batra, Oishi Deb, Emma Beharry, Cornelius Emde, Thomas Foster, Anna Gausen, Mar{\'i}a Grandury, Simeng Han, Valentin Hofmann, Lujain Ibrahim, Hazel Kim, Hannah~Rose Kirk, Fangru Lin, Gabrielle Kaili-May Liu, Lennart Luettgau, Jabez Magomere, Jonathan Rystr{\o}m, Anna Sotnikova, Yushi Yang, Yilun Zhao, Adel Bibi, Antoine Bosselut, Ronald Clark, Arman Cohan, Jakob~Nicolaus Foerster, Yarin Gal, Scott~A. Hale, Inioluwa~Deborah Raji, Christopher Summerfield, Philip H.~S. Torr, Cozmin Ududec, Luc Rocher, and Adam Mahdi.
\newblock Measuring what matters: Construct validity in large language model benchmarks, 2025.
\newblock URL \url{https://doi.org/10.48550/arXiv.2511.04703}.
\newblock Accepted to the NeurIPS 2025 Datasets and Benchmarks Track; title, acceptance status, and author list cross-checked against the Oxford RML project page and Oxford ORA record.

\bibitem[{Oxford Internet Institute}(2025)]{oxfordInternetInstitute2025BenchmarkPress}
{Oxford Internet Institute}.
\newblock Study identifies weaknesses in how {AI} systems are evaluated.
\newblock Press release, November 2025.
\newblock URL \url{https://www.oii.ox.ac.uk/news-events/study-identifies-weaknesses-in-how-ai-systems-are-evaluated/}.
\newblock Press release accompanying the benchmark-validity study; includes quoted claims about unclear definitions, weak methods, and misleading benchmark conclusions. Accessed 2026-03-21.

\bibitem[Sculley et~al.(2015)Sculley, Holt, Golovin, Davydov, Phillips, Ebner, Chaudhary, Young, Crespo, and Dennison]{sculleyEtAl2015HiddenTechnicalDebt}
D.~Sculley, Gary Holt, Daniel Golovin, Eugene Davydov, Todd Phillips, Dietmar Ebner, Vinay Chaudhary, Michael Young, Jean-Francois Crespo, and Dan Dennison.
\newblock Hidden technical debt in machine learning systems.
\newblock In \emph{Advances in Neural Information Processing Systems 28}, pages 2503--2511, 2015.
\newblock URL \url{https://papers.nips.cc/paper/5656-hidden-technical-debt-in-machine-learning-systems}.

\bibitem[Henderson et~al.(2018)Henderson, Islam, Bachman, Pineau, Precup, and Meger]{hendersonEtAl2018DeepRLMatters}
Peter Henderson, Riashat Islam, Philip Bachman, Joelle Pineau, Doina Precup, and David Meger.
\newblock Deep reinforcement learning that matters.
\newblock In \emph{Proceedings of the {AAAI} Conference on Artificial Intelligence}, volume~32, pages 3207--3214, 2018.
\newblock \doi{10.1609/AAAI.V32I1.11694}.
\newblock URL \url{https://doi.org/10.1609/AAAI.V32I1.11694}.

\bibitem[McGreivy and Hakim(2024)]{mcgreivyHakim2024WeakBaselines}
Nick McGreivy and Ammar Hakim.
\newblock Weak baselines and reporting biases lead to overoptimism in machine learning for fluid-related partial differential equations.
\newblock \emph{Nature Machine Intelligence}, 6\penalty0 (10):\penalty0 1256--1269, 2024.
\newblock \doi{10.1038/s42256-024-00897-5}.
\newblock URL \url{https://doi.org/10.1038/s42256-024-00897-5}.

\bibitem[Geirhos et~al.(2020)Geirhos, Jacobsen, Michaelis, Zemel, Brendel, Bethge, and Wichmann]{geirhosEtAl2020ShortcutLearning}
Robert Geirhos, J{\"o}rn-Henrik Jacobsen, Claudio Michaelis, Richard~S. Zemel, Wieland Brendel, Matthias Bethge, and Felix~A. Wichmann.
\newblock Shortcut learning in deep neural networks.
\newblock \emph{Nature Machine Intelligence}, 2\penalty0 (11):\penalty0 665--673, 2020.
\newblock \doi{10.1038/s42256-020-00257-z}.
\newblock URL \url{https://doi.org/10.1038/s42256-020-00257-z}.

\bibitem[Ulmer et~al.(2022)Ulmer, Hardmeier, and Frellsen]{ulmerHardmeierFrellsen2022DeepSignificance}
Dennis Ulmer, Christian Hardmeier, and Jes Frellsen.
\newblock deep-significance --- easy and meaningful statistical significance testing in the age of neural networks, 2022.
\newblock URL \url{https://doi.org/10.48550/arXiv.2204.06815}.
\newblock arXiv preprint; also listed as a contribution to the ML Evaluation Standards Workshop at ICLR 2022 in institutional repositories.

\bibitem[Jones et~al.(2024)Jones, Castro, De~Sousa~Ribeiro, Oktay, McCradden, and Glocker]{jonesEtAl2024CausalPerspective}
Charles Jones, Daniel~C. Castro, Fabio De~Sousa~Ribeiro, Ozan Oktay, Melissa McCradden, and Ben Glocker.
\newblock A causal perspective on dataset bias in machine learning for medical imaging.
\newblock \emph{Nature Machine Intelligence}, 6:\penalty0 138--146, 2024.
\newblock \doi{10.1038/s42256-024-00797-8}.
\newblock URL \url{https://doi.org/10.1038/s42256-024-00797-8}.

\bibitem[Broadbent and Grote(2022)]{broadbentGrote2022CanRobotsDoEpidemiology}
Alex Broadbent and Thomas Grote.
\newblock Can robots do epidemiology? machine learning, causal inference, and predicting the outcomes of public health interventions.
\newblock \emph{Philosophy \& Technology}, 35:\penalty0 14, 2022.
\newblock \doi{10.1007/s13347-022-00509-3}.
\newblock URL \url{https://doi.org/10.1007/s13347-022-00509-3}.
\newblock Springer presents this as volume 35, article number 14; issue and expanded page-range metadata vary across indexes.

\bibitem[Collins and Moons(2019)]{collinsMoons2019ReportingAIPredictionModels}
Gary~S. Collins and Karel G.~M. Moons.
\newblock Reporting of artificial intelligence prediction models.
\newblock \emph{The Lancet}, 393\penalty0 (10181):\penalty0 1577--1579, 2019.
\newblock \doi{10.1016/S0140-6736(19)30037-6}.
\newblock URL \url{https://doi.org/10.1016/S0140-6736(19)30037-6}.

\bibitem[Fuller-Wright(2024)]{fullerWright2024AISnakeOil}
Liz Fuller-Wright.
\newblock {``AI Snake Oil'': A Conversation with Princeton AI Experts Arvind Narayanan and Sayash Kapoor}.
\newblock Princeton University News, December 2024.
\newblock URL \url{https://www.princeton.edu/news/2024/12/18/ai-snake-oil-conversation-princeton-ai-experts-arvind-narayanan-and-sayash-kapoor}.
\newblock Interview/article quoting Narayanan and Kapoor on AI that does not work as advertised and predictive-AI systems not backed by scientific evidence. Accessed 2026-03-21.

\bibitem[Karpathy(2026)]{introKarpathyAutoResearch2026}
Andrej Karpathy.
\newblock Autoresearch.
\newblock GitHub repository, 2026.
\newblock URL \url{https://github.com/karpathy/autoresearch/blob/master/program.md}.
\newblock Repository documentation in \texttt{program.md}; accessed 2026-03-26.

\bibitem[Huang et~al.(2024)Huang, Vora, Liang, and Leskovec]{introHuangEtAlMLAgentBench2024}
Qian Huang, Jian Vora, Percy Liang, and Jure Leskovec.
\newblock Mlagentbench: Evaluating language agents on machine learning experimentation.
\newblock In \emph{Forty-first International Conference on Machine Learning}, 2024.
\newblock URL \url{https://openreview.net/forum?id=1Fs1LvjYQW}.

\bibitem[Lu et~al.(2024{\natexlab{a}})Lu, Lu, Lange, Foerster, Clune, and Ha]{introLuEtAlAIScientist2024}
Chris Lu, Cong Lu, Robert~Tjarko Lange, Jakob Foerster, Jeff Clune, and David Ha.
\newblock The {AI} scientist: Towards fully automated open-ended scientific discovery.
\newblock \emph{arXiv preprint arXiv:2408.06292}, 2024{\natexlab{a}}.
\newblock \doi{10.48550/arXiv.2408.06292}.
\newblock URL \url{https://arxiv.org/abs/2408.06292}.

\bibitem[Schmidgall et~al.(2025{\natexlab{a}})Schmidgall, Su, Wang, Sun, Wu, Yu, Liu, Moor, Liu, and Barsoum]{introSchmidgallEtAlAgentLaboratory2025}
Samuel Schmidgall, Yusheng Su, Ze~Wang, Ximeng Sun, Jialian Wu, Xiaodong Yu, Jiang Liu, Michael Moor, Zicheng Liu, and Emad Barsoum.
\newblock Agent laboratory: Using {LLM} agents as research assistants.
\newblock In Christos Christodoulopoulos, Tanmoy Chakraborty, Carolyn Rose, and Violet Peng, editors, \emph{Findings of the Association for Computational Linguistics: {EMNLP} 2025}, pages 5977--6043, Suzhou, China, November 2025{\natexlab{a}}. Association for Computational Linguistics.
\newblock ISBN 979-8-89176-335-7.
\newblock \doi{10.18653/v1/2025.findings-emnlp.320}.
\newblock URL \url{https://aclanthology.org/2025.findings-emnlp.320/}.

\bibitem[Chen et~al.(2025)Chen, Xiong, Lu, Han, Deng, He, Wu, Li, Liu, and Hooi]{introChenEtAlMLRBench2025}
Hui Chen, Miao Xiong, Yujie Lu, Wei Han, Ailin Deng, Yufei He, Jiaying Wu, Yibo Li, Yue Liu, and Bryan Hooi.
\newblock Mlr-bench: Evaluating {AI} agents on open-ended machine learning research.
\newblock \emph{arXiv preprint arXiv:2505.19955}, 2025.
\newblock \doi{10.48550/arXiv.2505.19955}.
\newblock URL \url{https://arxiv.org/abs/2505.19955}.

\bibitem[Ke et~al.(2015)Ke, Ferrara, Radicchi, and Flammini]{introKeEtAlSleepingBeauties2015}
Qing Ke, Emilio Ferrara, Filippo Radicchi, and Alessandro Flammini.
\newblock Defining and identifying sleeping beauties in science.
\newblock \emph{Proceedings of the National Academy of Sciences}, 112\penalty0 (24):\penalty0 7426--7431, 2015.
\newblock \doi{10.1073/pnas.1424329112}.
\newblock URL \url{https://www.pnas.org/doi/10.1073/pnas.1424329112}.

\bibitem[Hicks et~al.(2015)Hicks, Wouters, Waltman, de~Rijcke, and Rafols]{introHicksEtAlLeiden2015}
Diana Hicks, Paul Wouters, Ludo Waltman, Sarah de~Rijcke, and Ismael Rafols.
\newblock Bibliometrics: The leiden manifesto for research metrics.
\newblock \emph{Nature}, 520\penalty0 (7548):\penalty0 429--431, 2015.
\newblock \doi{10.1038/520429a}.
\newblock URL \url{https://www.nature.com/articles/520429a}.

\bibitem[Wilsdon et~al.(2015)Wilsdon, Allen, Belfiore, Campbell, Curry, Hill, Jones, Hill, Kain, Johnson, Kerridge, Tinkler, Thelwall, Wouters, and Viney]{introWilsdonEtAlMetricTide2015}
James Wilsdon, Liz Allen, Eleonora Belfiore, Philip Campbell, Stephen Curry, Steven Hill, Richard Jones, Jude Hill, Roger Kain, Ben Johnson, Simon Kerridge, Jane Tinkler, Mike Thelwall, Paul Wouters, and Ian Viney.
\newblock The metric tide: Report of the independent review of the role of metrics in research assessment and management.
\newblock Technical report, Higher Education Funding Council for England, London, 2015.
\newblock URL \url{https://hdl.handle.net/10779/uos.23418680}.

\bibitem[Fire and Guestrin(2019)]{introFireGuestrinGoodhart2019}
Michael Fire and Carlos Guestrin.
\newblock Over-optimization of academic publishing metrics: Observing goodhart's law in action.
\newblock \emph{GigaScience}, 8\penalty0 (6):\penalty0 giz053, 2019.
\newblock \doi{10.1093/gigascience/giz053}.
\newblock URL \url{https://doi.org/10.1093/gigascience/giz053}.

\bibitem[Du et~al.(2024)Du, Li, Torralba, Tenenbaum, and Mordatch]{du2024improving}
Yilun Du, Shuang Li, Antonio Torralba, Joshua~B Tenenbaum, and Igor Mordatch.
\newblock Improving factuality and reasoning in language models through multiagent debate.
\newblock In \emph{Forty-first international conference on machine learning}, 2024.

\bibitem[Funkquist et~al.(2022)Funkquist, Kuznetsov, Hou, and Gurevych]{funkquist2023citebench}
Martin Funkquist, Ilia Kuznetsov, Yufang Hou, and Iryna Gurevych.
\newblock Citebench: A benchmark for scientific citation text generation, 2022.
\newblock URL \url{https://arxiv.org/abs/2212.09577}.
\newblock Using the arXiv submission year; later bibliographic records may surface under 2023 metadata updates.

\bibitem[Li et~al.(2024)Li, Chen, Liu, Yu, and Wen]{li2024chatcite}
Yutong Li, Lu~Chen, Aiwei Liu, Kai Yu, and Lijie Wen.
\newblock Chatcite: {LLM} agent with human workflow guidance for comparative literature summary, 2024.
\newblock URL \url{https://arxiv.org/abs/2403.02574}.

\bibitem[Wang et~al.(2025)Wang, Ma, Nie, Zeng, Lyu, Zhang, Schneider, Lu, Yue, and Chen]{wang2025scholarcopilot}
Yubo Wang, Xueguang Ma, Ping Nie, Huaye Zeng, Zhiheng Lyu, Yuxuan Zhang, Benjamin Schneider, Yi~Lu, Xiang Yue, and Wenhu Chen.
\newblock Scholarcopilot: Training large language models for academic writing with accurate citations, 2025.
\newblock URL \url{https://arxiv.org/abs/2504.00824}.

\bibitem[Bao et~al.(2025)Bao, Nayeem, Rafiei, and Zhang]{bao2025surveygen}
Tong Bao, Mir~Tafseer Nayeem, Davood Rafiei, and Chengzhi Zhang.
\newblock Surveygen: Quality-aware scientific survey generation with large language models, 2025.
\newblock URL \url{https://arxiv.org/abs/2508.17647}.

\bibitem[Wen et~al.(2024)Wen, Wei, Lin, Wang, Liang, and Wan]{wen2024overleafcopilot}
Haomin Wen, Zhenjie Wei, Yan Lin, Jiyuan Wang, Yuxuan Liang, and Huaiyu Wan.
\newblock Overleafcopilot: Empowering academic writing in {Overleaf} with large language models, 2024.
\newblock URL \url{https://arxiv.org/abs/2403.09733}.

\bibitem[Hou et~al.(2025)Hou, Lin, Chen, Gong, and He]{hou2025paperdebugger}
Junyi Hou, Huikai~Andre Lin, Nuo Chen, Yiwei Gong, and Bingsheng He.
\newblock Paperdebugger: A plugin-based multi-agent system for in-editor academic writing, review, and editing, 2025.
\newblock URL \url{https://arxiv.org/abs/2512.02589}.

\bibitem[Ifargan et~al.(2025)Ifargan, Hafner, Kern, Alcalay, and Kishony]{ifargan2025datatopaper}
Tal Ifargan, Lukas Hafner, Maor Kern, Ori Alcalay, and Roy Kishony.
\newblock Autonomous {LLM}-driven research --- from data to human-verifiable research papers.
\newblock \emph{NEJM AI}, 2\penalty0 (1), 2025.
\newblock \doi{10.1056/AIoa2400555}.
\newblock URL \url{https://ai.nejm.org/doi/10.1056/AIoa2400555}.

\bibitem[Lu et~al.(2024{\natexlab{b}})Lu, Lu, Lange, Foerster, Clune, and Ha]{lu2024aiscientist}
Chris Lu, Cong Lu, Robert~Tjarko Lange, Jakob Foerster, Jeff Clune, and David Ha.
\newblock The {AI} scientist: Towards fully automated open-ended scientific discovery, 2024{\natexlab{b}}.
\newblock URL \url{https://arxiv.org/abs/2408.06292}.

\bibitem[Yamada et~al.(2025)Yamada, Lange, Lu, Hu, Lu, Foerster, Clune, and Ha]{yamada2025aiscientistv2}
Yutaro Yamada, Robert~Tjarko Lange, Cong Lu, Shengran Hu, Chris Lu, Jakob Foerster, Jeff Clune, and David Ha.
\newblock The {AI} scientist-v2: Workshop-level automated scientific discovery via agentic tree search, 2025.
\newblock URL \url{https://arxiv.org/abs/2504.08066}.

\bibitem[Schmidgall et~al.(2025{\natexlab{b}})Schmidgall, Su, Wang, Sun, Wu, Yu, Liu, Moor, Liu, and Barsoum]{schmidgall2025agentlaboratory}
Samuel Schmidgall, Yusheng Su, Ze~Wang, Ximeng Sun, Jialian Wu, Xiaodong Yu, Jiang Liu, Michael Moor, Zicheng Liu, and Emad Barsoum.
\newblock Agent laboratory: Using {LLM} agents as research assistants.
\newblock In \emph{Findings of the Association for Computational Linguistics: EMNLP 2025}, pages 5977--6043. Association for Computational Linguistics, 2025{\natexlab{b}}.
\newblock \doi{10.18653/v1/2025.findings-emnlp.320}.
\newblock URL \url{https://aclanthology.org/2025.findings-emnlp.320/}.

\bibitem[Weng et~al.(2024)Weng, Zhu, Bao, Zhang, Wang, Zhang, and Yang]{weng2025cycleresearcher}
Yixuan Weng, Minjun Zhu, Guangsheng Bao, Hongbo Zhang, Jindong Wang, Yue Zhang, and Linyi Yang.
\newblock Cycleresearcher: Improving automated research via automated review, 2024.
\newblock URL \url{https://arxiv.org/abs/2411.00816}.
\newblock First submitted in 2024; later revised in 2025.

\bibitem[Tang et~al.(2025)Tang, Xia, Li, and Huang]{tang2025airesearcher}
Jiabin Tang, Lianghao Xia, Zhonghang Li, and Chao Huang.
\newblock {AI}-researcher: Autonomous scientific innovation, 2025.
\newblock URL \url{https://arxiv.org/abs/2505.18705}.

\bibitem[Schmidgall and Moor(2025)]{schmidgall2025agentrxiv}
Samuel Schmidgall and Michael Moor.
\newblock Agentrxiv: Towards collaborative autonomous research, 2025.
\newblock URL \url{https://arxiv.org/abs/2503.18102}.

\bibitem[Li et~al.(2025)Li, Ren, Pan, Yan, Li, Bergemann, and Yang]{li2025freephdlabor}
Ed~Li, Junyu Ren, Xintian Pan, Cat Yan, Chuanhao Li, Dirk Bergemann, and Zhuoran Yang.
\newblock Build your personalized research group: A multiagent framework for continual and interactive science automation, 2025.
\newblock URL \url{https://arxiv.org/abs/2510.15624}.

\bibitem[Gottweis et~al.(2025)Gottweis, Weng, Daryin, Tu, Palepu, Sirkovic, Myaskovsky, Weissenberger, Rong, Tanno, Saab, Popovici, Blum, Zhang, Chou, Hassidim, Gokturk, Vahdat, Kohli, Matias, Carroll, Kulkarni, Tomasev, Guan, Dhillon, Vaishnav, Lee, Costa, Penad{\'e}s, Peltz, Xu, Pawlosky, Karthikesalingam, and Natarajan]{gottweis2025aicoscientist}
Juraj Gottweis, Wei-Hung Weng, Alexander Daryin, Tao Tu, Anil Palepu, Petar Sirkovic, Artiom Myaskovsky, Felix Weissenberger, Keran Rong, Ryutaro Tanno, Khaled Saab, Dan Popovici, Jacob Blum, Fan Zhang, Katherine Chou, Avinatan Hassidim, Burak Gokturk, Amin Vahdat, Pushmeet Kohli, Yossi Matias, Andrew Carroll, Kavita Kulkarni, Nenad Tomasev, Yuan Guan, Vikram Dhillon, Eeshit~Dhaval Vaishnav, Byron Lee, Tiago R.~D. Costa, Jos{\'e}~R. Penad{\'e}s, Gary Peltz, Yunhan Xu, Annalisa Pawlosky, Alan Karthikesalingam, and Vivek Natarajan.
\newblock Towards an {AI} co-scientist, 2025.
\newblock URL \url{https://arxiv.org/abs/2502.18864}.

\bibitem[{InternAgent Team} et~al.(2025){InternAgent Team}, Zhang, Feng, Yan, Yuan, Ma, Hu, Yu, He, Huang, Hou, Nie, Wang, Liu, Peng, Ye, Zhou, Zhang, Wang, Zhang, Li, Tu, Yue, Ouyang, Zhou, and Bai]{internagentteam2025internagent}
{InternAgent Team}, Bo~Zhang, Shiyang Feng, Xiangchao Yan, Jiakang Yuan, Runmin Ma, Yusong Hu, Zhiyin Yu, Xiaohan He, Songtao Huang, Shaowei Hou, Zheng Nie, Zhilong Wang, Jinyao Liu, Tianshuo Peng, Peng Ye, Dongzhan Zhou, Shufei Zhang, Xiaosong Wang, Yilan Zhang, Meng Li, Zhongying Tu, Xiangyu Yue, Wangli Ouyang, Bowen Zhou, and Lei Bai.
\newblock Internagent: When agent becomes the scientist --- building closed-loop system from hypothesis to verification, 2025.
\newblock URL \url{https://arxiv.org/abs/2505.16938}.

\bibitem[Feng et~al.(2026)Feng, Ma, Yan, Fan, Hu, Huang, Zhang, Cao, Peng, Yuan, Guo, Zhong, Du, Wang, Shi, Zhou, He, Yu, Yu, Zheng, Wu, Liu, Zhang, Hou, Li, Jiang, Lou, Wang, Wang, Wang, Xu, Deng, Liu, Wang, Zhang, Ling, Zhang, Wang, Zheng, Huang, Sun, Hu, Ye, Song, Wang, He, Liu, Li, Hou, Chen, Yue, He, Lin, Zhou, Zhang, and Bai]{feng2026internagent15}
Shiyang Feng, Runmin Ma, Xiangchao Yan, Yue Fan, Yusong Hu, Songtao Huang, Shuaiyu Zhang, Zongsheng Cao, Tianshuo Peng, Jiakang Yuan, Zijie Guo, Zhijie Zhong, Shangheng Du, Weida Wang, Jinxin Shi, Yuhao Zhou, Xiaohan He, Zhiyin Yu, Fangchen Yu, Qihao Zheng, Jiamin Wu, Mianxin Liu, Chi Zhang, Shaowei Hou, Shuya Li, Yankai Jiang, Wenjie Lou, Lilong Wang, Zifu Wang, Jiong Wang, Wanghan Xu, Yue Deng, Dongrui Liu, Yiheng Wang, Wenlong Zhang, Fenghua Ling, Shufei Zhang, Xiaosong Wang, Shuangjia Zheng, Xun Huang, Siqi Sun, Shuyue Hu, Peng Ye, Chunfeng Song, Bin Wang, Conghui He, Yihao Liu, Xin Li, Qibin Hou, Tao Chen, Xiangyu Yue, Liang He, Dahua Lin, Bowen Zhou, Bo~Zhang, and Lei Bai.
\newblock Internagent-1.5: A unified agentic framework for long-horizon autonomous scientific discovery, 2026.
\newblock URL \url{https://arxiv.org/abs/2602.08990}.

\bibitem[Novikov et~al.(2025{\natexlab{b}})Novikov, V{\~u}, Eisenberger, Dupont, Huang, Wagner, Shirobokov, Kozlovskii, Ruiz, Mehrabian, Kumar, See, Chaudhuri, Holland, Davies, Nowozin, Kohli, and Balog]{novikov2025alphaevolve}
Alexander Novikov, Ng{\^a}n V{\~u}, Marvin Eisenberger, Emilien Dupont, Po-Sen Huang, Adam~Zsolt Wagner, Sergey Shirobokov, Borislav Kozlovskii, Francisco J.~R. Ruiz, Abbas Mehrabian, M.~Pawan Kumar, Abigail See, Swarat Chaudhuri, George Holland, Alex Davies, Sebastian Nowozin, Pushmeet Kohli, and Matej Balog.
\newblock Alphaevolve: A coding agent for scientific and algorithmic discovery, 2025{\natexlab{b}}.
\newblock URL \url{https://arxiv.org/abs/2506.13131}.

\bibitem[Georgiev et~al.(2025{\natexlab{b}})Georgiev, G{\'o}mez-Serrano, Tao, and Wagner]{georgiev2025mathematicalexploration}
Bogdan Georgiev, Javier G{\'o}mez-Serrano, Terence Tao, and Adam~Zsolt Wagner.
\newblock Mathematical exploration and discovery at scale, 2025{\natexlab{b}}.
\newblock URL \url{https://arxiv.org/abs/2511.02864}.

\bibitem[Fan et~al.(2026)Fan, Zhang, Zheng, Chen, Niu, Huang, Lin, and Huang]{fan2026deepinnovator}
Tianyu Fan, Fengji Zhang, Yuxiang Zheng, Bei Chen, Xinyao Niu, Chengen Huang, Junyang Lin, and Chao Huang.
\newblock Deepinnovator: Triggering the innovative capabilities of {LLM}s, 2026.
\newblock URL \url{https://arxiv.org/abs/2602.18920}.

\bibitem[Weidener et~al.(2026)Weidener, Brki{\'c}, Lee, Karlsson, Noessler, and Kohlhaas]{weidener2026clawdlab}
Lukas Weidener, Marko Brki{\'c}, Phillip Lee, Martin Karlsson, Kevin Noessler, and Paul Kohlhaas.
\newblock From agent-only social networks to autonomous scientific research: Lessons from {OpenClaw} and {Moltbook}, and the architecture of {ClawdLab} and {Beach.Science}, 2026.
\newblock URL \url{https://arxiv.org/abs/2602.19810}.

\bibitem[Lo et~al.(2020)Lo, Wang, Neumann, Kinney, and Weld]{lo2020s2orc}
Kyle Lo, Lucy~Lu Wang, Mark Neumann, Rodney Kinney, and Daniel Weld.
\newblock S2orc: The semantic scholar open research corpus.
\newblock In \emph{Proceedings of the 58th Annual Meeting of the Association for Computational Linguistics}, pages 4969--4983. Association for Computational Linguistics, 2020.
\newblock \doi{10.18653/v1/2020.acl-main.447}.
\newblock URL \url{https://aclanthology.org/2020.acl-main.447/}.

\bibitem[Priem et~al.(2022)Priem, Piwowar, and Orr]{priem2022openalex}
Jason Priem, Heather Piwowar, and Richard Orr.
\newblock Openalex: A fully-open index of scholarly works, authors, venues, institutions, and concepts, 2022.
\newblock URL \url{https://arxiv.org/abs/2205.01833}.

\bibitem[Gao et~al.(2023)Gao, Yen, Yu, and Chen]{gao2023alce}
Tianyu Gao, Howard Yen, Jiatong Yu, and Danqi Chen.
\newblock Enabling large language models to generate text with citations.
\newblock In \emph{Proceedings of the 2023 Conference on Empirical Methods in Natural Language Processing}, pages 6465--6488. Association for Computational Linguistics, 2023.
\newblock \doi{10.18653/v1/2023.emnlp-main.398}.
\newblock URL \url{https://aclanthology.org/2023.emnlp-main.398/}.

\bibitem[Skarlinski et~al.(2024)Skarlinski, Cox, Laurent, Braza, Hinks, Hammerling, Ponnapati, Rodriques, and White]{skarlinski2024paperqa2}
Michael~D. Skarlinski, Sam Cox, Jon~M. Laurent, James~D. Braza, Michaela Hinks, Michael~J. Hammerling, Manvitha Ponnapati, Samuel~G. Rodriques, and Andrew~D. White.
\newblock Language agents achieve superhuman synthesis of scientific knowledge, 2024.
\newblock URL \url{https://arxiv.org/abs/2409.13740}.

\bibitem[{Elicit}(n.d.)]{elicit2026site}
{Elicit}.
\newblock Elicit: {AI} for scientific research, n.d.
\newblock URL \url{https://orion.elicit.com/}.
\newblock Undated product site; accessed 2026-03-21.

\bibitem[{OpenClaw}(n.d.)]{openclaw2026docs}
{OpenClaw}.
\newblock Openclaw documentation, n.d.
\newblock URL \url{https://docs.openclaw.ai/}.
\newblock Undated documentation site; accessed 2026-03-21.

\bibitem[{ClawdLab}(n.d.)]{clawdlab2026site}
{ClawdLab}.
\newblock Clawdlab, n.d.
\newblock URL \url{https://www.clawdlab.xyz/}.
\newblock Undated project site; accessed 2026-03-21.

\bibitem[{Molecule Protocol}(n.d.)]{sciencebeach2026repo}
{Molecule Protocol}.
\newblock science.beach.
\newblock GitHub repository, n.d.
\newblock URL \url{https://github.com/moleculeprotocol/science.beach}.
\newblock Undated repository citation; accessed 2026-03-21.

\bibitem[Yang et~al.(2023)Yang, Swope, Gu, Chalamala, Song, Yu, Godil, Prenger, and Anandkumar]{yang2023leandojo}
Kaiyu Yang, Aidan~M. Swope, Alex Gu, Rahul Chalamala, Peiyang Song, Shixing Yu, Saad Godil, Ryan Prenger, and Anima Anandkumar.
\newblock Leandojo: Theorem proving with retrieval-augmented language models, 2023.
\newblock URL \url{https://arxiv.org/abs/2306.15626}.

\bibitem[Zheng et~al.(2021)Zheng, Han, and Polu]{zheng2021minif2f}
Kunhao Zheng, Jesse~Michael Han, and Stanislas Polu.
\newblock minif2f: A cross-system benchmark for formal olympiad-level mathematics, 2021.
\newblock URL \url{https://arxiv.org/abs/2109.00110}.

\bibitem[Trinh et~al.(2024)Trinh, Wu, Le, He, and Luong]{trinh2024alphageometry}
Trieu~H. Trinh, Yuhuai Wu, Quoc~V. Le, He~He, and Thang Luong.
\newblock Solving olympiad geometry without human demonstrations.
\newblock \emph{Nature}, 625\penalty0 (7995):\penalty0 476--482, 2024.
\newblock \doi{10.1038/s41586-023-06747-5}.
\newblock URL \url{https://www.nature.com/articles/s41586-023-06747-5}.

\bibitem[{AlphaProof and AlphaGeometry teams}(2024)]{alphaproof2024blog}
{AlphaProof and AlphaGeometry teams}.
\newblock {AI} achieves silver-medal standard solving international mathematical olympiad problems.
\newblock Google DeepMind blog, July 2024.
\newblock URL \url{https://deepmind.google/blog/ai-solves-imo-problems-at-silver-medal-level/}.
\newblock Published 2024-07-25; accessed 2026-03-21.

\bibitem[Mialon et~al.(2024)Mialon, Fourrier, Swift, Wolf, LeCun, and Scialom]{mialon2024gaia}
Gr{\'e}goire Mialon, Cl{\'e}mentine Fourrier, Craig Swift, Thomas Wolf, Yann LeCun, and Thomas Scialom.
\newblock {GAIA}: A benchmark for general {AI} assistants.
\newblock In \emph{The Twelfth International Conference on Learning Representations}, 2024.
\newblock URL \url{https://openreview.net/forum?id=fibxvahvs3}.

\bibitem[Rein et~al.(2023)Rein, Hou, Stickland, Petty, Pang, Dirani, Michael, and Bowman]{rein2023gpqa}
David Rein, Betty~Li Hou, Asa~Cooper Stickland, Jackson Petty, Richard~Yuanzhe Pang, Julien Dirani, Julian Michael, and Samuel~R. Bowman.
\newblock {GPQA}: A graduate-level google-proof {Q\&A} benchmark, 2023.
\newblock URL \url{https://arxiv.org/abs/2311.12022}.

\bibitem[Phan et~al.(2025)Phan, Gatti, Han, Li, Hu, Zhang, Zhang, Shaaban, Ling, Shi, et~al.]{phan2025humanity}
Long Phan, Alice Gatti, Ziwen Han, Nathaniel Li, Josephina Hu, Hugh Zhang, Chen Bo~Calvin Zhang, Mohamed Shaaban, John Ling, Sean Shi, et~al.
\newblock Humanity's last exam, 2025.
\newblock URL \url{https://arxiv.org/abs/2501.14249}.

\bibitem[{OpenAI}(2025{\natexlab{b}})]{openai2025frontierscience}
{OpenAI}.
\newblock Evaluating {AI}'s ability to perform scientific research tasks.
\newblock OpenAI research publication, 2025{\natexlab{b}}.
\newblock URL \url{https://openai.com/index/frontierscience/}.
\newblock Published 2025-12-16; accessed 2026-03-21.

\end{thebibliography}
\bibliographystyle{unsrtnat}

\appendix

\section{Full Architecture Reference}
\label{sec:architecture_reference}

This appendix provides a self-contained technical reference for the \textbf{pAI}\texttt{/MSc} architecture: the execution graph, agent inventory, track-level details, optional rigor modules, campaign orchestration, and the project-level code layout. It consolidates material that was spread across the system-overview, design-choices, and implementation sections of an earlier draft.

\subsection{Full pipeline graph description}
\label{sec:app_pipeline}

At the core of the release is a fixed single-run LangGraph workflow. ``Fixed'' refers to the execution topology: the graph always traverses the same high-level stages and routing logic, even though the generated content is stochastic.

The graph is organized into six phases.

\paragraph{Phase 1: Discovery and feasibility screening.}
The run begins with \texttt{persona\_council}, followed by \texttt{literature\_\allowbreak review\_agent}, a literature-review gate, \texttt{brainstorm\_\allowbreak agent}, and \texttt{formalize\_\allowbreak goals\_agent}. This phase grounds the task in prior work, checks feasibility, and converts a high-level objective into a concrete technical plan.

\paragraph{Phase 2: Track planning and routing.}
Planning writes a decomposition into theory, experiment, or mixed tracks, recorded in \texttt{paper\_workspace/\allowbreak track\_decomposition.json}. The track router uses this contract to decide which execution branches to launch.

\paragraph{Phase 3: Parallel technical execution.}
When theory work is selected and theorem-oriented modules are enabled, the graph runs the theory path through mathematical literature review, proposal, proving, rigorous verification, empirical verification, and proof transcription. When empirical work is selected, it runs experiment literature review, design, execution, verification, and transcription. Both tracks can run independently or in parallel.

\paragraph{Phase 4: Completion verification and results formalization.}
Track outputs are merged via \texttt{track\_merge} and passed to \texttt{verify\_\allowbreak completion}, which routes in three directions: \emph{complete} (proceed to results synthesis), \emph{incomplete} (return to goal formalization), or \emph{rethink} (return to brainstorming). If the run proceeds, \texttt{formalize\_\allowbreak results\_agent} produces a structured assessment in \texttt{results\_\allowbreak assessment.pdf}.

\paragraph{Phase 5: Internal consistency and follow-up.}
A duality check is applied after results formalization. Failure triggers follow-up literature work and renewed planning; success advances to paper production. The file \texttt{followup\_decision.json} captures the routing outcome.

\paragraph{Phase 6: Paper production and editorial quality assurance.}
The pipeline runs \texttt{resource\_\allowbreak preparation\_agent}, \texttt{writeup\_agent}, \texttt{proofreading\_\allowbreak agent}, \texttt{reviewer\_agent}, and \texttt{validation\_gate}. Validation failure routes back to writeup for revision before a final success state.

\paragraph{Loopbacks.}
A failed literature-feasibility screen routes back to \texttt{persona\_\allowbreak council}; \texttt{verify\_\allowbreak completion} can send the run back to \texttt{formalize\_\allowbreak goals\_agent} or to \texttt{brainstorm\_agent}; a failed duality gate triggers follow-up literature work; and a failed validation gate returns control to writeup.

\paragraph{Human steering surfaces.}
The system exposes: (i) an initial task specification passed to the launcher; (ii) live steering through a TCP control socket and HTTP API during execution; (iii) resume from a named stage; (iv) context injection via an \texttt{inputs/} directory; and (v) campaign-level control when the optional campaign engine is used.

\subsection{Agent inventory and specialist roles}
\label{sec:app_agents}

The current release decomposes work into the following specialist agents, grouped by phase.

\begin{table}[h]
\centering
\footnotesize
\setlength{\tabcolsep}{3pt}
\begin{tabular}{|L{1.8cm}|L{4.4cm}|L{6.2cm}|}
\hline
\textbf{Phase} & \textbf{Agent} & \textbf{Role} \\
\hline
Discovery & \texttt{persona\_council} & Three-perspective debate (practitioner, theorist, narrative) to evaluate direction \\
\hline
Discovery & \texttt{literature\_review\allowbreak\_agent} & Literature grounding and feasibility gate \\
\hline
Planning & \texttt{brainstorm\_agent} & Convert research prompt to operational ideas \\
\hline
Planning & \texttt{formalize\_goals\allowbreak\_agent} & Structured goal state, milestone definition \\
\hline
Theory & \texttt{math\_literature\allowbreak\_agent} & Mathematical literature review for the theory track \\
\hline
Theory & \texttt{math\_proposer\allowbreak\_agent} & Propose claims, lemmas, and theorem statements \\
\hline
Theory & \texttt{math\_prover\_agent} & Construct proof drafts \\
\hline
Theory & \texttt{math\_rigorous\allowbreak\_verifier\_agent} & Rigorous verification of proof drafts \\
\hline
Theory & \texttt{math\_empirical\allowbreak\_verifier\_agent} & Empirical/numerical checks on mathematical claims \\
\hline
Theory & \texttt{proof\_transcription\allowbreak\_agent} & Transcribe accepted proofs into paper workspace \\
\hline
Experiment & \texttt{experiment\_literature\allowbreak\_agent} & Empirical literature review and baseline identification \\
\hline
Experiment & \texttt{experiment\_design\allowbreak\_agent} & Experimental design specification \\
\hline
Experiment & \texttt{experimentation\_agent} & Execute experiments (local subprocess, Docker, or SLURM) \\
\hline
Experiment & \texttt{experiment\allowbreak\_verification\_agent} & Post-hoc verification of experimental results \\
\hline
Experiment & \texttt{experiment\allowbreak\_transcription\_agent} & Transcribe results into paper workspace \\
\hline
Synthesis & \texttt{track\_merge} & Merge theory and experiment outputs \\
\hline
Synthesis & \texttt{verify\_completion} & Three-way routing: complete / incomplete / rethink \\
\hline
Synthesis & \texttt{formalize\_results\allowbreak\_agent} & Structured results assessment \\
\hline
Editorial & \texttt{resource\_preparation\allowbreak\_agent} & Prepare assets, bibliography, resource inventory \\
\hline
Editorial & \texttt{writeup\_agent} & Manuscript generation (markdown or \LaTeX{}) \\
\hline
Editorial & \texttt{proofreading\_agent} & Copyediting and revision artifacts \\
\hline
Editorial & \texttt{reviewer\_agent} & Internal review with scoring \\
\hline
Editorial & \texttt{validation\_gate} & Final artifact-contract and review-threshold check \\
\hline
\end{tabular}
\caption{Agent inventory in the current \textbf{pAI}\texttt{/MSc} release, grouped by pipeline phase.}
\label{tab:agent_inventory}
\end{table}

\subsection{Theory track details}
\label{sec:app_theory}

The theorem-oriented path is optional and enabled only when the planner selects theory work and math agents are available (\texttt{--enable-math-agents}). The path includes six agents: \texttt{math\_literature\_\allowbreak agent}, \texttt{math\_proposer\_\allowbreak agent}, \texttt{math\_prover\_\allowbreak agent}, \texttt{math\_rigorous\_\allowbreak verifier\_agent}, \texttt{math\_empirical\_\allowbreak verifier\_agent}, and \texttt{proof\_transcription\_\allowbreak agent}.

The corresponding artifact surface is explicit: the math workspace stores a claim graph in \texttt{claim\_graph.json}, proof files under \texttt{proofs/}, verification traces under \texttt{checks/}, and a reusable lemma store in \texttt{lemma\_\allowbreak library.md}. The mathematical path is treated as a workspace with claims, dependencies, and verification traces rather than as one long free-form proof draft.

These agents provide structure for theorem-oriented work and for integrating accepted claims into the paper, but they do not establish mathematical truth by oracle. The rigorous and empirical verifiers are supportive checks, not substitutes for expert proof review.

\subsection{Experiment track details}
\label{sec:app_experiment}

When empirical work is selected, the graph runs \texttt{experiment\_\allowbreak literature\_agent}, \texttt{experiment\_\allowbreak design\_agent}, \texttt{experimentation\_\allowbreak agent}, \texttt{experiment\_\allowbreak verification\_agent}, and \texttt{experiment\_\allowbreak transcription\_agent}.

This path writes its own workspace: \texttt{experiment\_\allowbreak literature.md}, \texttt{experiment\_\allowbreak baselines.json}, \texttt{experiment\_\allowbreak design.json}, \texttt{experiment\_\allowbreak rationale.md}, \texttt{execution\_log.json}, \texttt{verification\_\allowbreak report.md}, and \texttt{verification\_\allowbreak results.json}. When experiment transcription runs, the paper workspace additionally receives \texttt{experiment\_\allowbreak report.tex} and \texttt{experiment\_\allowbreak report.pdf}.

Experiment execution is delegated to an external execution layer. In the current release this includes AI-Scientist-v2 integration and optional SLURM submission. Verification occurs \emph{after} execution and serves as a post-hoc assessment, not a pre-execution code safety guarantee.

\subsection{Counsel protocol specification}
\label{sec:app_counsel}

When \texttt{--enable-counsel} is active, selected specialist stages become a multi-model protocol with sandboxing, debate, synthesis, and artifact promotion.

\paragraph{Model pool.} In the default configuration, the debate pool includes Anthropic Claude Opus 4.6, OpenAI GPT-5.4, and Google Gemini 3 Pro Preview, with Claude Sonnet 4.6 as the synthesis model. These choices are configurable through the model configuration file.

\paragraph{Protocol.} Each model first works in an isolated sandbox copy of the workspace, producing an independent candidate output. Models then critique the candidate solutions over several debate rounds. A synthesis model selects and promotes a single authoritative output back to the main workspace.

\paragraph{Target stages.} Counsel targets stages where disagreement is informative: literature review, experiment design, and manuscript writing.

\paragraph{Cost profile.} Counsel increases per-stage cost substantially but changes the failure mode from ``one model silently committed to one line of reasoning'' to ``multiple models exposed and critiqued alternative drafts before promotion.''

\subsection{Tree search algorithm}
\label{sec:app_tree_search}

When \texttt{--enable-tree-search} is active, the linear proof stage in the theory track is replaced by a DAG-layered best-first controller that explores multiple proof strategies in parallel.

\paragraph{Algorithm.}
\begin{enumerate}
    \item Frontier claims are selected from the claim graph.
    \item Alternative proof strategies are generated for each frontier claim.
    \item Branches are scored using promise, graph impact, cost efficiency, depth, and diversity.
    \item Top candidates are executed in forked workspaces.
    \item Failed branches can spawn debugging children up to a configurable depth.
    \item Low-scoring branches are pruned.
\end{enumerate}

\paragraph{Artifacts.} The search state is persisted in \texttt{tree\_search\_state.json} and branch-level workspaces are stored under \texttt{tree\_branches/}, making the search process inspectable rather than hidden inside one monolithic prover call.

\subsection{Campaign orchestration}
\label{sec:app_campaign}

The single-run graph is the primary execution unit. On top of it, the project provides an optional campaign layer driven by heartbeat scripts, cron, or OpenClaw.

Campaigns repeatedly launch or resume runs, validate stage artifacts, distill stage memory, track budget, attempt autonomous repair when artifacts are missing, and notify the user. Campaign state is recorded explicitly, and circuit breakers prevent runaway budget consumption.

The launcher supports full-workspace resume, resume from a named stage, and addition of context files under \texttt{inputs/}. During execution, the interaction layer exposes both a TCP control socket and an HTTP steering API, allowing pause, instruction injection, and status inspection without manually reconstructing context.

\subsection{Mathematical reasoning pipeline}
\label{sec:app_math_pipeline}

The theory track implements a structured mathematical reasoning pipeline rather than a single free-form proof attempt:

\begin{enumerate}
    \item \textbf{Mathematical literature review} (\texttt{math\_literature\_agent}): surveys existing results relevant to the theory claims.
    \item \textbf{Claim proposal} (\texttt{math\_proposer\_agent}): proposes claims, lemmas, and theorem statements, populating the claim graph.
    \item \textbf{Proof construction} (\texttt{math\_prover\_agent}): builds proof drafts for claims in the graph. When tree search is enabled, this stage expands into parallel branch exploration.
    \item \textbf{Rigorous verification} (\texttt{math\_rigorous\_\allowbreak verifier\_agent}): checks proof drafts for logical gaps, missing cases, and dependency errors.
    \item \textbf{Empirical verification} (\texttt{math\_empirical\_\allowbreak verifier\_agent}): runs numerical or constructive checks against mathematical claims.
    \item \textbf{Transcription} (\texttt{proof\_transcription\_agent}): promotes accepted proofs into the paper workspace as LaTeX-ready theorem environments.
\end{enumerate}

The claim graph (\texttt{claim\_graph.json}) tracks claim status (proposed, proved, verified, transcribed, failed) and inter-claim dependencies. The lemma library (\texttt{lemma\_library.md}) accumulates reusable intermediate results across proof attempts.

\subsection{Project structure and module map}
\label{sec:app_project_structure}

The repository is organized around a thin launcher and a core package. The top-level entry point \texttt{launch\_multiagent.py} is lightweight; most workflow logic lives inside the \texttt{consortium} package.

\begin{table}[h]
\centering
\footnotesize
\setlength{\tabcolsep}{3pt}
\begin{tabular}{|L{2.8cm}|L{4.0cm}|L{5.4cm}|}
\hline
\textbf{Component} & \textbf{Location} & \textbf{Role} \\
\hline
Launcher and run lifecycle & \texttt{launch\_multi\-agent.py}, \texttt{consortium/\allowbreak runner.py} & CLI entry point, config loading, workspace creation, checkpoint setup, execution lifecycle \\
\hline
Workflow topology and state & \texttt{consortium/\allowbreak graph.py}, \texttt{consortium/\allowbreak state.py}, \texttt{consortium/\allowbreak workflow\_utils.py} & LangGraph definition, track routing, gates, shared validation helpers, state schema \\
\hline
Model and counsel layer & \texttt{consortium/\allowbreak models.py}, \texttt{consortium/\allowbreak utils.py}, \texttt{consortium/\allowbreak counsel.py} & Model registry, provider mappings, multi-model sandbox/\allowbreak debate/\allowbreak synthesis \\
\hline
Specialist agents & \texttt{consortium/\allowbreak agents/} & Discovery agents, theory agents, experiment agents, merge, writeup, proofreading, reviewer \\
\hline
Domain toolkits & \texttt{consortium/\allowbreak toolkits/*} & Search, experimentation, writeup, math, filesystem, ideation, communication \\
\hline
Validation and supervision & \texttt{consortium/\allowbreak supervision/} & Artifact validation, review checks, traceability checks \\
\hline
Optional advanced modules & \texttt{consortium/\allowbreak tree\_search/}, \texttt{consortium/\allowbreak campaign/}, \texttt{consortium/\allowbreak interaction/}, \texttt{consortium/\allowbreak external\_tools/} & Tree search, campaign engine, steering APIs, experiment execution integrations \\
\hline
\end{tabular}
\caption{Implementation boundary. \textbf{pAI}\texttt{/MSc} is the project-level system; \texttt{consortium} is the package that implements the workflow.}
\label{tab:module_map_appendix}
\end{table}

\subsection{Run directory layout}
\label{sec:app_run_layout}

Each run writes to a timestamped directory under \texttt{results/consortium\_\allowbreak <timestamp>/}. The run directory is the main interface through which the system exposes its internal state.

\begin{itemize}
    \item \textbf{Run root.} Top-level deliverables: \texttt{final\_paper.tex}, optionally \texttt{final\_paper.pdf}, plus \texttt{run\_token\_usage.json}, budget files, \texttt{checkpoints.db}, inter-agent messages, and memory backup.
    \item \textbf{Paper workspace} (\texttt{paper\_workspace/}). Literature review, research plan, risk register, track decomposition, results assessment, follow-up decision, follow-up literature notes, experiment report transcription, resource inventory, and bibliography.
    \item \textbf{Experiment workspace} (\texttt{experiment\_workspace/}). Baselines, design specification, rationale, execution log, and verification reports.
    \item \textbf{Experiment runs} (\texttt{experiment\_runs/<uuid>/}). Concrete experiment executions launched by the external execution layer.
    \item \textbf{Math workspace} (\texttt{math\_workspace/}). Claim graph, proof drafts, checks, and lemma library (when theorem-oriented agents are enabled).
    \item \textbf{Counsel sandboxes} (\texttt{counsel\_sandboxes/}). Per-agent, per-model sandbox outputs before synthesis (when counsel is enabled).
    \item \textbf{Tree-search state} (\texttt{tree\_search\_state.json} and \texttt{tree\_branches/}). Branch-level search state and forked workspaces (when tree search is enabled).
\end{itemize}

\section{Extended Related Work}
\label{sec:extended_related_work}

The literature closest to \textbf{pAI}\texttt{/MSc} does not form a single clean lineage. It spans manuscript-local writing assistants, literature-review and survey systems, manuscript-first research-to-paper agents, broader autonomous discovery systems, and open multi-agent scientific platforms. We find it more useful to compare these systems by three questions than by raw agent count: \emph{what scholarly artifact is produced}, \emph{what substrate the system can actually act on}, and \emph{what verification loop constrains error accumulation}. Under that lens, \textbf{pAI}\texttt{/MSc} is best understood as a manuscript-first, artifact-driven research-to-paper system for quantitative research, rather than as a generic autonomous-science platform.

\subsection{Historical precursors: scholarly infrastructure, citation grounding, and editor-native assistance}

The pre-agent history of research automation is largely a history of stronger anchoring. Before multi-agent research workflows became feasible, the main bottlenecks were access to machine-readable scholarly corpora, citation graphs, and interfaces for grounding generated text in evidence. At the corpus level, \citet{lo2020s2orc} introduced S2ORC, a large-scale structured corpus that made citation-aware retrieval and document-level text mining practical at scale. At the catalog and graph level, \citet{priem2022openalex} introduced OpenAlex as a fully open index of works, authors, institutions, venues, and concepts, giving later systems a programmable substrate for literature discovery, citation traversal, and bibliometric filtering. These resources matter because many later ``research agents'' inherit their apparent capability from the quality of the scholarly substrate underneath them.

A first wave of LLM-era work then focused on citation-grounded text generation rather than full research automation. \citet{gao2023alce} introduced ALCE, a benchmark and evaluation framework for end-to-end generation of text with citations, emphasizing fluency, correctness, and citation quality rather than stylistic quality alone. \citet{funkquist2023citebench} similarly proposed CiteBench as a benchmark for citation text generation, helping expose the gap between merely \emph{having} citations and citing evidence that actually supports a claim. On top of this evaluation infrastructure, project-level systems such as ChatCite \citep{li2024chatcite}, ScholarCopilot \citep{wang2025scholarcopilot}, SurveyGen \citep{bao2025surveygen}, PaperQA2 \citep{skarlinski2024paperqa2}, and Elicit \citep{elicit2026site} pushed increasingly far into literature synthesis, comparative summary, citation-grounded writing, and systematic-review-style workflows. These systems are highly relevant to \textbf{pAI}\texttt{/MSc} because they establish an important baseline lesson: in scholarly writing, retrieval quality and claim--evidence linkage often matter more than surface fluency.

A parallel line binds generation directly to live manuscript state rather than to a free-form chat session. OverleafCopilot \citep{wen2024overleafcopilot} embeds assistance in an Overleaf/\LaTeX{} workflow, while PaperDebugger \citep{hou2025paperdebugger} adds plugin-based multi-agent review and patching in-editor. This editor-native family is adjacent rather than identical to hypothesis-to-paper systems such as ours, but it highlights a design principle that we also adopt: manuscript state, intermediate artifacts, and revision traces are valuable external objects and should not be collapsed into ephemeral conversational context.

\subsection{Manuscript-first research-to-paper agents}

The closest prior line to \textbf{pAI}\texttt{/MSc} is the family of systems that explicitly target end-to-end or near end-to-end manuscript production. Within this family, a useful internal distinction is between \emph{provenance-first} systems, which emphasize traceability from evidence to prose, and \emph{exploration-first} systems, which emphasize autonomous idea and experiment search before paper synthesis.

Among provenance-first systems, data-to-paper is the clearest reference point. \citet{ifargan2025datatopaper} describe a stepwise framework that starts from structured data, generates hypotheses, produces analyses and code, and then writes a human-verifiable manuscript with explicit backward traceability from claims to executed steps. The main affinity with \textbf{pAI}\texttt{/MSc} is therefore not simply ``end-to-end research,'' but the insistence that a manuscript should be supported by inspectable intermediate artifacts rather than by polished prose alone.

A second cluster emphasizes broader autonomous pipeline coverage. The AI Scientist \citep{lu2024aiscientist} popularized the idea of a repeated loop in which a system generates research ideas, writes code, runs experiments, drafts a full paper, and then critiques its own output through a reviewer-style stage. AI Scientist-v2 \citep{yamada2025aiscientistv2} extends this exploration-first pattern through agentic tree search and stronger workshop-level paper generation. Agent Laboratory \citep{schmidgall2025agentlaboratory} structures the workflow into literature review, experimentation, and report writing, while explicitly showing that human feedback at stage boundaries materially improves outcomes. AI-Researcher \citep{tang2025airesearcher} frames autonomous scientific innovation itself as the core task, and CycleResearcher \citep{weng2025cycleresearcher} explicitly couples automated generation to automated review in a revision loop. AgentRxiv \citep{schmidgall2025agentrxiv} adds a collaborative preprint-server abstraction on top of these laboratory-style agents, enabling cross-run accumulation and retrieval of prior reports rather than treating every run as an isolated research episode.

freephdlabor \citep{li2025freephdlabor} is especially relevant because it departs from rigid fixed pipelines and instead argues for highly customizable, fully dynamic workflows determined by real-time agent reasoning. It is therefore much closer than earlier systems to a ``research operating system'' for multi-agent science. Relative to freephdlabor, however, \textbf{pAI}\texttt{/MSc} intentionally remains more constrained: our emphasis is not maximum architectural freedom, but a tighter artifact contract, clearer theory--experiment coupling, and stronger paper-facing validation gates.

Taken together, these systems establish the manuscript-first research-agent frontier into which \textbf{pAI}\texttt{/MSc} enters. Our system is closest to this family in artifact contract, but it differs in emphasis. Relative to exploration-first systems, we place more weight on bounded workflow control, explicit intermediate artifacts, and the coordination of theory and experiments under a single paper trajectory. Relative to more data-first systems, we begin from a human-specified hypothesis and support both mathematical and empirical development of that idea.

\subsection{Broader autonomous scientific discovery systems}

A nearby but distinct literature optimizes for autonomous scientific discovery more broadly rather than for manuscript production as the primary output. The difference matters: these systems are often stronger on ideation, search breadth, or domain-specific discovery, yet weaker on paper-facing artifact contracts and manuscript governance.

At one end of this spectrum is AI co-scientist. \citet{gottweis2025aicoscientist} describe a multi-agent system that proposes, debates, and iteratively refines scientific hypotheses under scientist guidance, with biomedical validation examples. The center of gravity here is hypothesis generation and research reasoning, not manuscript production. InternAgent \citep{internagentteam2025internagent} and InternAgent-1.5 \citep{feng2026internagent15} push further toward long-horizon, closed-loop autonomous discovery across multiple scientific domains, with architectures explicitly organized around generation, verification, and evolution. DeepInnovator \citep{fan2026deepinnovator} addresses a different bottleneck still: not orchestration of the full pipeline, but training models for innovative idea generation through structured research memories and iterative next-idea prediction. In mathematics and algorithmic discovery, AlphaEvolve \citep{novikov2025alphaevolve} uses evaluator-guided coding agents for scientific and algorithmic discovery, while Mathematical Exploration and Discovery at Scale \citep{georgiev2025mathematicalexploration} studies large-scale machine-assisted mathematical exploration more directly.

These systems are important to our positioning because they enlarge the meaning of ``research automation.'' They suggest that the space is not exhausted by paper-writing systems, and that one can build powerful discovery engines without centering the final manuscript artifact. \textbf{pAI}\texttt{/MSc}, by contrast, is not intended as a general theory of autonomous scientific discovery. It is narrower: a system whose organizing question is how far one can push a hypothesis-to-paper workflow while keeping human steering sparse and scientific artifacts explicit.

\subsection{Open multi-agent scientific platforms and decentralized research commons}

A newer line of work moves away from both bounded single-run pipelines and monolithic laboratory agents toward persistent, open, multi-agent scientific ecosystems. The most important reference here is the OpenClaw/Moltbook ecosystem analyzed by \citet{weidener2026clawdlab}. OpenClaw itself is not a manuscript-first research agent in the narrow sense; its official documentation presents it as a self-hosted gateway and plugin platform for AI agents across messaging channels and skills \citep{openclaw2026docs}. What makes it relevant to research automation is that it became a substrate on top of which scientific-agent ecosystems were rapidly built.

Within that ecosystem, ClawdLab and Beach.Science instantiate two distinct architectural responses. The ClawdLab paper of \citet{weidener2026clawdlab} presents ClawdLab as a structured laboratory platform with hard role restrictions, PI-led governance, explicit adversarial critique, and externally verifiable evidence requirements. The public ClawdLab site \citep{clawdlab2026site} reinforces this emphasis through specialized roles, permanent reports, and cryptographically auditable logs. Beach.Science, by contrast, is presented both in \citet{weidener2026clawdlab} and in the public repository \citep{sciencebeach2026repo} as an open research commons in which heterogeneous agents and humans exchange, evaluate, and refine hypotheses in a more decentralized environment.

This platform-oriented line is important because it introduces design questions that bounded workflow papers often understate: agent identity, governance, auditability, reward design, role restrictions, and persistent public interaction. Relative to ClawdLab or Beach.Science, \textbf{pAI}\texttt{/MSc} is deliberately less open-ended and less ecosystem-oriented. It is designed as a governed research-to-paper pipeline rather than as a general-purpose research commons. But these platforms are still highly relevant because they reveal what breaks when one relaxes workflow constraints, and because they show that future scientific-agent systems may need governance and auditability mechanisms that exceed what isolated pipeline papers currently provide.

\subsection{Verification substrates for machine learning theory and applied mathematics}

Because \textbf{pAI}\texttt{/MSc} is built with a current emphasis on machine learning theory and adjacent quantitative fields, it should also be positioned relative to work on formal mathematical verification, not only relative to manuscript agents. Quantitative disciplines differ from many applied domains in that they sometimes admit verification loops much stronger than human critique alone.

LeanDojo \citep{yang2023leandojo} is a central enabling platform in this regard: it turns Lean repositories into programmatically accessible training and evaluation environments for theorem proving, with structured proof states, premises, and proof traces. miniF2F \citep{zheng2021minif2f} provides a cross-system benchmark for Olympiad-style formal mathematics, allowing direct comparison of theorem provers across formal systems. At a more public frontier, AlphaGeometry \citep{trinh2024alphageometry} demonstrated that synthetic-data-driven neuro-symbolic reasoning can reach olympiad-level geometry performance, and AlphaProof \citep{alphaproof2024blog} brought reinforcement-learning-based formal reasoning into the IMO setting through Lean-verified proofs.

These systems are not research-to-paper agents in the ordinary sense. However, they are directly relevant to theorem-heavy workflows because they illustrate a stronger class of verification loop: when full or partial formalization is feasible, the most meaningful verifier is not another critic agent but a compiler-like checker. This is one reason \textbf{pAI}\texttt{/MSc} treats theorem-oriented reasoning as an optional rigor-enhancing extension rather than as a cosmetic extra.

\subsection{Evaluation context: capability benchmarks versus manuscript benchmarks}

Finally, several recent benchmarks matter for positioning capability claims, even though they are not manuscript-quality benchmarks. GAIA \citep{mialon2024gaia} evaluates general AI assistants on real-world, tool-using tasks. GPQA \citep{rein2023gpqa} probes graduate-level ``Google-proof'' scientific question answering. Humanity's Last Exam \citep{phan2025humanity} seeks a broad frontier academic benchmark that remains difficult even as models saturate older tests. FrontierScience \citep{openai2025frontierscience} evaluates expert-level scientific reasoning on olympiad-style and research-style scientific subtasks.

These benchmarks are useful, but they should not be confused with evaluation of manuscript-centered research systems. A system can score well on GAIA, GPQA, or FrontierScience and still fail to produce a traceable, coherent, properly evidenced paper. Conversely, a system can be strong on structured manuscript production while being narrow in open-ended scientific reasoning. This is precisely why we position \textbf{pAI}\texttt{/MSc} not by a single scalar ``research capability'' score, but by its artifact contract, its grounding substrate, and its verification design.

\section{Operational Reference}
\label{sec:operational_reference}

\subsection{Cost and runtime summary}
\label{sec:app_cost}

Table~\ref{tab:runtime_cost} in the main body summarizes cost and runtime envelopes for representative configurations. For convenience, the key ranges are: quickstart runs cost \$2--10 in 15--40 min; base pipeline with LaTeX/PDF costs \$10--40 in 30--90 min; math-agent runs cost \$20--60 in 60--150 min; counsel mode costs \$50--200 in 2--5 hrs; tree search without counsel costs \$60--180 in 2--4 hrs; tree search with counsel costs \$200--600 in 4--10 hrs; and full paper campaigns cost \$100--400 in 6--12 hrs. Actual values vary with task scope, model choice, revision loops, and experiment intensity.

The configuration system resolves model settings through a precedence order: built-in defaults, then \texttt{.llm\_config.yaml}, then explicit CLI overrides. Budget tracking uses workspace-local files: \texttt{budget\_state.json}, \texttt{budget\_ledger.jsonl}, and \texttt{run\_token\_usage.json}.

\subsection{Experiment safety model}
\label{sec:app_safety}

The experiment path is deliberately usable, but its safety model remains limited. Table~\ref{tab:safety_model_appendix} summarizes what is and is not in place.

\begin{table}[h]
\centering
\footnotesize
\setlength{\tabcolsep}{3pt}
\begin{tabular}{|L{2.6cm}|L{4.6cm}|L{5.0cm}|}
\hline
\textbf{Aspect} & \textbf{What is in place} & \textbf{What is not in place} \\
\hline
Process separation & Experiments run in dedicated \texttt{experiment\_runs/\allowbreak <uuid>/} directories; local execution wrapped in a child process & Local experiments run as the same OS user when not containerized \\
\hline
Time control & Local timeout via \texttt{CONSORTIUM\allowbreak\_EXPERIMENT\allowbreak\_TIMEOUT}; SLURM wall-time in cluster mode & No general-purpose per-experiment CPU or memory isolation in local mode \\
\hline
Environment isolation & Optional Docker for whole-pipeline execution; optional SLURM for cluster enforcement & No mandatory per-experiment container, namespace, or chroot \\
\hline
Network and resource control & Budget cap on API spend & No network isolation; no local cgroup/\allowbreak ulimit enforcement \\
\hline
Code safety & Input-schema validation and post-hoc experiment verification & No human code-review gate before execution of AI-generated code \\
\hline
\end{tabular}
\caption{Experiment-execution safety model. The release includes pragmatic controls but not strong sandbox guarantees.}
\label{tab:safety_model_appendix}
\end{table}

The correct interpretation: the system is suitable for controlled research environments where the operator accepts these risks and uses Docker or SLURM when appropriate, but it should not be described as a hardened execution sandbox.

\subsection{Submission checklist}
\label{sec:app_checklist}

Before submitting any \textbf{pAI}\texttt{/MSc}-produced manuscript to a venue (NeurIPS, ICML, ICLR, or similar), a human owner should at minimum:

\begin{enumerate}
    \item Verify novelty and related-work positioning against recent proceedings, not only arXiv.
    \item Confirm that baselines are appropriate, figures are not misleading, and at least one key experiment has been rerun independently.
    \item Spot-check citations and bibliography entries, including retraction status and faithfulness of attributed claims.
    \item Manually revise high-stakes sections: abstract, ethics or broader-impact discussion, and venue-specific formatting or anonymization details.
    \item Compile the manuscript locally and obtain at least one domain-expert read before submission.
    \item Check venue-specific requirements: page limits, supplementary material format, code-availability statements, and reproducibility checklists.
\end{enumerate}

This checklist is part of the scientific boundary of the system. \textbf{pAI}\texttt{/MSc} reduces orchestration burden but does not remove the need for expert ownership of the final claims.

\section{Evaluation Framework}
\label{sec:evaluation_framework}

This appendix specifies the evaluation structure for \textbf{pAI}\texttt{/MSc}: the core evaluation questions, the metrics taxonomy, and the known failure modes. The current release should be evaluated primarily as a \emph{workflow and artifact system}, not as a benchmarked scientific autopilot.

\subsection{Evaluation questions}
\label{sec:app_eval_questions}

For the present release, the central evaluation questions are:

\begin{itemize}
    \item \textbf{Q1 -- Structural completion.} Does the system reliably execute the fixed workflow graph and materialize the required stage artifacts under different operating modes?
    \item \textbf{Q2 -- Editorial completion.} Under stricter settings, does the run reach a paper-facing state in which the manuscript artifacts, review artifacts, and optional PDF outputs are all present and structurally valid?
    \item \textbf{Q3 -- Auditability.} Do the emitted files make it possible for a human supervisor to understand what the system planned, what it executed, what it decided to follow up on, and where additional verification is still required?
    \item \textbf{Q4 -- Operational efficiency.} What runtime and spend envelopes are induced by the major execution modes, and how do optional modules such as counsel and tree search change that profile?
    \item \textbf{Q5 -- Scientific quality.} Which aspects of output quality can already be instrumented automatically, and which still require external auditing, reruns, or expert review?
\end{itemize}

These questions deliberately separate \emph{operational success} (Q1--Q4, already measurable from run outputs) from \emph{scientific success} (Q5, only partially observable in the current release).

\subsection{Metrics taxonomy}
\label{sec:app_metrics}

Evaluation metrics fall into three families with different evidentiary status.

\paragraph{Structural metrics.}
Directly observable from workspace and validator outputs:
artifact completion rate,
strict editorial-gate pass rate,
PDF compile rate when required,
resume success after interruption,
branch- and stage-level completion statistics.

\paragraph{Operational metrics.}
Already instrumented by run-level files (\texttt{run\_token\_usage.json}, \texttt{budget\_ledger.jsonl}, completion summaries):
total runtime,
per-stage runtime,
total spend,
token counts by model,
number of reviewer or rebuttal loops,
branch counts in tree-search mode.

\paragraph{Scientific-audit metrics.}
Require a separate evaluation protocol beyond the current package:
number of human steering interventions,
citation-faithfulness spot checks,
independent reruns of experiments,
proof review outcomes,
external expert judgments of manuscript quality.

Metrics in the first two families support release characterization today; metrics in the third family are the target for a future benchmark paper.

\subsection{Failure modes and ablation axes}
\label{sec:app_failures}

Table~\ref{tab:failure_modes_appendix} catalogs the recurring failure modes, current mitigations, and unresolved boundaries.

\begin{table}[h]
\centering
\footnotesize
\setlength{\tabcolsep}{3pt}
\begin{tabular}{|L{2.8cm}|L{4.6cm}|L{4.8cm}|}
\hline
\textbf{Failure mode} & \textbf{Current mitigation} & \textbf{What remains unresolved} \\
\hline
Late-stage churn and overlong finishing loops & Validation gate, reviewer loop, configurable rebuttal iterations, milestone gates, campaign circuit breakers & A run can still spend substantial time polishing a weak draft instead of improving core content \\
\hline
Theory/\allowbreak experiment drift & Explicit \texttt{track\_decomposition\allowbreak.json}, shared merge stage, completion verification, duality check, follow-up routing & Global coherence depends on intermediate reasoning quality, not only route structure \\
\hline
Weak manuscript quality in fully autonomous mode & Artifact enforcement, proofreading, reviewer stage, optional counsel, optional math agents & The system stops short of claiming submission-ready quality without human verification \\
\hline
Citation brittleness & Literature review stage, bibliography generation, submission checklist requiring manual spot checks & No internal oracle for citation truthfulness or complete related-work coverage \\
\hline
\LaTeX{} fragility & Writeup reflection loop, fail-fast prerequisite checks, optional markdown mode & Complex papers may require manual debugging and local compilation \\
\hline
Budget explosion in higher-rigor modes & Budget caps, token ledgers, cost summaries, counsel/\allowbreak tree-search controls & Counsel and tree search remain expensive; quality gains need empirical justification \\
\hline
Experiment safety and correctness risk & Subprocess timeout, optional Docker, optional SLURM, post-hoc verification & No per-experiment sandbox, no network isolation, no pre-execution code review gate \\
\hline
\end{tabular}
\caption{Release-stage failure modes, current mitigations, and unresolved boundaries.}
\label{tab:failure_modes_appendix}
\end{table}

\paragraph{Suggested ablation axes.} The most informative future studies would isolate: (i) the effect of counsel on editorial quality; (ii) the effect of tree search on theorem-track productivity; (iii) the effect of strict artifact enforcement on completion rates; and (iv) the effect of campaign repair on long-horizon task success.

\end{document}